\documentclass{article}

\usepackage[preprint]{neurips_2026}

\usepackage[utf8]{inputenc}
\usepackage[T1]{fontenc}
\usepackage{hyperref}
\usepackage{url}
\usepackage{booktabs}
\usepackage{amsfonts}
\usepackage{amsmath}
\usepackage{nicefrac}
\usepackage{microtype}
\usepackage{xcolor}
\usepackage{graphicx}
\usepackage{subcaption}
\usepackage{enumitem}
\usepackage{flafter}
\usepackage{float}
\usepackage{placeins}
\usepackage{array}
\usepackage{tabularx}
\newcolumntype{R}[1]{>{\raggedright\arraybackslash\hspace{0pt}}p{#1}}
\hypersetup{hidelinks}

\title{What Must a Fairness Audit Report When Demographic Data Is Incomplete?}

\newcommand{\RAC}{\texttt{RAC1P}}
\newcommand{\SEX}{\texttt{SEX}}
\newcommand{\INT}{\texttt{RAC1P$\times$SEX}}

\newcommand{\pp}{pp}
\newcommand{\ADS}{Audit Disclosure Standard}
\newcommand{\appendixtablegap}{\par\vspace{0.9em plus 0.25em minus 0.15em}}
\newcommand{\appendixpagegap}{\FloatBarrier\appendixtablegap}

\definecolor{adsnum}{HTML}{9A8078}
\definecolor{adsevidence}{HTML}{56504E}
\newcommand{\adspanelfirst}[2]{%
  \noalign{\vspace{4pt}}%
  \multicolumn{3}{@{}l@{}}{\textbf{\textsc{#1}\hspace{0.35em}\textsc{#2}}}\\
  \noalign{\vspace{3pt}}}
\newcommand{\adspanel}[2]{%
  \noalign{\vspace{8pt}}%
  \multicolumn{3}{@{}l@{}}{\textbf{\textsc{#1}\hspace{0.35em}\textsc{#2}}}\\
  \noalign{\vspace{3pt}}}
\newcommand{\adsrow}[4]{{\textcolor{adsnum}{\scriptsize #1}\hspace{0.4em}#2} & #3 & \textcolor{adsevidence}{#4}\\[1.4pt]}

\author{%
  Yash Vardhan Tomar \\
  Purdue University\\
  \texttt{tomar4@purdue.edu} \\
}

\begin{document}

\maketitle

\begin{abstract}
Fairness audits are a key component of responsible machine-learning deployment. Yet what such an
audit must disclose, when the protected labels it depends on are incomplete, remains unsettled. In
this work, we focused on the rates a fairness audit publishes and on what an oversight reader needs
beside them. We paired every published rate with a matched baseline drawn from the same audit, one
hiding protected labels and one varying only the run seed. Across ACS/Folktables tasks, missingness
settings that kept some protected labels moved the selected mitigation less than an ordinary rerun
did. At zero protected-label access, candidates collapsed to empirical risk minimization, so the
apparent exception there reflected the candidate set's composition. Equalized-odds threshold optimization most often regressed an
intersectional subgroup, but that rate fell back to its baseline once we kept only the configurations
an auditor would accept. Any accuracy it lost fell on the population as heavily as on the worst-off
cell, so the mechanism is levelling down. The one effect that survived was a change in which cell is
worst-off. Overall, our results highlight that a published
audit rate should be reported with the baseline needed to interpret it, the candidate set it came
from, and its intersectional effects, before it is treated as evidence about a deployed model.
\end{abstract}

\section{Introduction}

Fairness audits are increasingly produced as part of institutional oversight, where they inform
deployment decisions. Each audit is a document that compares candidate mitigations, measures utility
alongside group disparities, and recommends one method under a pre-specified fairness--utility rule.
Agencies procuring risk-scoring tools and firms filing under sectoral anti-discrimination rules
produce much the same document. Whether an oversight reader can trust it turns on the evidence it
contains for interpreting its own conclusion, much more than on how sophisticated the mitigation was.

Audits depend on protected attributes, meaning labels such as race or sex that identify the groups
whose error rates and outcomes are being compared, and these labels enter the process at several
points. Training-time and post-processing mitigations may use them to fit or adjust a model, and
evaluation uses them again for group-level and intersectional metrics. When they are missing for some
of the people assessed, the audit can no longer say who benefited, who was harmed, or whether the
disparity a mitigation targeted actually moved.

Most institutions that run audits do so without complete demographic data, whether because they never
collected it, because the people assessed decline to provide it, or because their records are
inconsistent. In some settings the missingness is itself group-dependent. A large body of work addresses auditing or training when
protected attributes are unobserved, inferred from proxies, or noisy
\citep{chen2019fairness,kallus2020assessing,lahoti2020fairness,
wang2020robust,ghosh2023noisy,kenfack2024survey}, and a smaller one shows how missing values distort
fairness claims \citep{martinez2021missing,min2025inequitable}. Mitigation conclusions also shift with
task, model class, and metric \citep{chen2023comprehensive,wang2025inferred}. Demographic information
is therefore part of the measurement itself, because how labels are observed or lost changes both the
disparities it reports and the mitigation it selects.

\paragraph{Two failure modes.}
One failure is under-reporting who was harmed. This happens whenever a model looks acceptable on every
protected axis the report covers while harming a structured subgroup. This failure is known as fairness
gerrymandering \citep{kearns2018preventing}, and later work develops it in worst-case subgroup
comparisons \citep{ghosh2021intersectional,chen2024multiple}. Audits that report race and sex separately can
also record a real improvement achieved by degrading the model for everyone, a pattern
\citet{mittelstadt2023unfairness} call levelling down, which neither the single-axis gap nor the
aggregate accuracy figure reveals. The opposite failure is over-reading instability. When labels are hidden
and the selected method changes, a raw flip rate gives no scale for judging that change. Even with complete labels,
different random seeds, near-equivalent candidates, and selection tie-breaks produce different
recommendations
\citep{black2022model,long2023individual,amir2021seeds,coston2021goodmodels,
watsondaniels2024predictive,herasymuk2025tradeoffs,dai2025intentional}. So without a seed baseline,
missingness looks more disruptive than the evidence supports, and scarce oversight attention goes to
noise. Both failures are properties of the report, so specifying what a report must
contain is enough to fix them. In this work, we therefore address what a fairness audit must
disclose, when the protected labels it depends on are incomplete, before an oversight reader can tell
substantive subgroup harm from routine audit noise. There are two reasons to settle this. First,
without a baseline for a published rate, a reader cannot tell whether a changed recommendation
reflects the labels that went missing or the ordinary movement of the audit. Second, without the
per-cell view behind an aggregate, a report can record a fairness gain that was bought by degrading
the model for everyone.

\paragraph{Approach and contributions.}
We instantiate a full audit workflow on four Folktables American Community Survey (ACS) prediction
tasks \citep{ding2021retiring}, with a candidate set containing empirical risk minimization (ERM),
feature mixing, reweighing \citep{kamiran2012data}, protected-group mixing, and equalized-odds
threshold optimization \citep{hardt2016equality}. We evaluate race (\RAC{}), sex (\SEX{}), and their
intersection (\INT{}) under missing completely at random (MCAR) availability sweeps and matched
missing not at random (MNAR) ablations, with a random-forest validation run. Two matched comparisons
inside the same audit cell separate the effects an oversight reader cares about from the noise
already in the process. One compares a missing-label run against the complete-label oracle, the other
two complete-label runs at different seeds. We contribute:

\begin{enumerate}[topsep=2pt,itemsep=1pt,parsep=0pt,leftmargin=1.5em]
  \item \textbf{The \ADS{}} (\autoref{tab:standard}), nine reporting fields that make an audit's own
    rates interpretable, grouped by the question each answers, each tied to a quantified misreading it
    prevents (\S\ref{sec:standard}).
  \item \textbf{Five calibration cases from one benchmark.} Each is a rate that reverses or shrinks
    once we supply the missing context. A recommendation-change rate falls below its own rerun
    baseline, a no-label rate is inflated by candidates with identical predictions, a
    subgroup-regression rate is computed mostly over configurations the audit would reject, and a
    worst-cell identity rate is 73 raw but 15 points once its null is subtracted
    (\S\ref{sec:hidden}, \S\ref{sec:calibration}).
  \item \textbf{A null-calibrated levelling-down result.} Equalized-odds post-processing degrades the
    population at least as much as the worst-off intersectional cell, so the subgroup harm it produces
    is levelling down. Non-threshold mitigations stay below the null, which makes this a specific
    post-processing failure an audit can be required to check for (\S\ref{sec:hidden}).
\end{enumerate}

\section{Audit setup and diagnostics}
\label{sec:setup}

In this section we specify the audit whose published rates the rest of the paper calibrates, and the
two diagnostics we compute from it.

\paragraph{Tasks, splits, and protected views.}
We build the audit on ACS/Folktables because it gives us repeated splits over the same population,
protected-label structure, and cells large enough to evaluate intersectionally. We use four binary
tasks: ACSIncome, ACSEmployment, ACSPublicCoverage, and
ACSMobility \citep{ding2021retiring}, drawn from ACS person files for 2018, 2019, 2021, and 2022
across all 50 states, with rolling temporal splits and leave-one-state-out geographic splits over 25
held-out states, giving 28 splits per task and $4\times28\times5=560$ task/split/seed clusters. A protected view is an
attribute axis used to form audit groups, and we use \RAC{}, \SEX{}, and \INT{}. Candidates are ERM,
feature mixing, reweighing, protected-group mixing, and equalized-odds threshold optimization.
Feature mixing interpolates same-label rows from different state-year contexts. Protected-group
mixing applies the same routine with observed protected-group labels in place of context, so it
collapses to ERM whenever no groups are observed. Threshold optimization fits a base model, then learns
group-specific equalized-odds thresholds on a 25\% validation split \citep{hardt2016equality}.
Primary results use logistic regression, with a random-forest validation run.
Appendix~\ref{app:config} gives hyperparameters, row caps, and seeds.

\paragraph{Missingness scenarios.}
In the complete-label scenario $r_0$ all protected labels are revealed. We always retain the oracle ACS labels for evaluation, method selection, and seed-null calibration. Missingness hides labels only
from training and prediction-time mitigation, so the design measures a change in the
recommendation, not added uncertainty about the evaluation groups. MCAR scenarios reveal labels at 80, 60, 50, 40, 20, 10, and
0\%. Matched MNAR scenarios use 50, 20, and 10\% targets, with a majority \INT{} cell receiving nominal
availability and all other cells one quarter of it. This quarter share creates a clear asymmetry in who loses
labels without erasing nearly every non-majority label at the sparsest settings, where too few
retained cells would survive to evaluate. We record a run seed for every audit because
seed movement is part of the calibration target.

\paragraph{Metrics.}
We compute all audit metrics on oracle-labeled test rows once the retained groups are fixed. Write
$Y\in\{0,1\}$ for the true label and $\widehat{Y}$ for a test prediction. A task/split/seed cluster is $u=(t,q,s)$, where $t$
indexes one of the four prediction tasks, $q$ a train/test split, and $s$ a run seed, giving 560
clusters. A candidate method is $m$ and a missingness scenario is $r$. These three arguments are
suppressed in the definitions below. Let $n_a(u)$ be oracle-labeled test support for group $a$ and
$\mathcal{G}_A(u)=\{a:n_a(u)\ge 25\}$ the retained groups for protected view $A$, so that $\max_a$ and
$\min_a$ range over $\mathcal{G}_A(u)$. Note that the 25-row rule keeps very small cells from being treated as stable estimates. Demographic parity (DP) is the largest difference in positive prediction rates
across retained groups,
$\mathrm{DP}_A=\max_a \Pr(\widehat{Y}=1\mid A=a)-\min_a \Pr(\widehat{Y}=1\mid A=a)$. Writing the
group true- and false-positive rates as $\mathrm{TPR}_{A,a}=\Pr(\widehat{Y}=1\mid Y=1,A=a)$ and
$\mathrm{FPR}_{A,a}=\Pr(\widehat{Y}=1\mid Y=0,A=a)$, equalized odds (EO) is the larger of the two
gaps,
\begin{equation}
  \mathrm{EO}_A=\max\Big\{\max_a \mathrm{TPR}_{A,a}-\min_a \mathrm{TPR}_{A,a},\;\;
  \max_a \mathrm{FPR}_{A,a}-\min_a \mathrm{FPR}_{A,a}\Big\}.
  \label{eq:eo-gap}
\end{equation}
Accuracy $\mathrm{Acc}$ is the fraction of oracle-labeled test rows classified correctly, and
worst-group accuracy (WGA), $\mathrm{WGA}_A$, is the minimum per-group accuracy over retained
groups, or how badly the system works for the group it works worst for. Note that WGA is a minimum,
so two audits can report the same value while the cell attaining it differs between them. We return
to this in \S\ref{sec:hidden}, where it changes how the statistic should be read.

\paragraph{The selector.}
A missingness scenario is $r\in\{r_0\}\cup\mathcal{R}$, where $\mathcal{R}$ collects the ten
scenarios that hide labels, candidate methods are $m\in\mathcal{M}$, and
selection criteria $c\in\mathcal{C}$ cover accuracy-constrained EO, DP, WGA, and intersectional EO.
Let $L_c(m,u,r)$ be the criterion loss, with smaller values preferred. It is the maximum of $\mathrm{EO}_A$ over the
three views for accuracy-constrained EO, the maximum of $\mathrm{DP}_A$ for DP, the negated minimum
of $\mathrm{WGA}_A$ for WGA, and $\mathrm{EO}_{\INT{}}$ for intersectional EO. Let
$\mathcal{E}_c(u,r)$ be the eligible methods. For accuracy-constrained EO these are the ones within 1.0
percentage point of ERM accuracy in the matched cluster and scenario, so
$\mathrm{Acc}(m,u,r)\ge \mathrm{Acc}(\mathrm{ERM},u,r)-0.01$; for other criteria, all of
$\mathcal{M}$. The one-point tolerance treats small accuracy differences as practically comparable
while stopping a fairness-only method from being selected after a large utility loss. We resolve exact ties by a fixed alphabetical rank $\rho(m)$, and selection is lexicographic:
\begin{equation}
  \widehat{m}_c(u,r)
  =
  \operatorname*{arg\,min}_{m\in \mathcal{E}_c(u,r)}
  \big(L_c(m,u,r),-\mathrm{Acc}(m,u,r),\rho(m)\big).
  \label{eq:selection}
\end{equation}
In other words, the audit takes the method with the smallest criterion loss, breaks near-ties on
accuracy, and falls back on name order whenever two candidates are indistinguishable. The tie breaker is
part of the audit specification because a recommendation rule must still return one method when
candidates become empirically identical, which is exactly what happens at 0\%
availability. The party that assembles these candidates and applies this rule is referred to as the auditor.

\paragraph{The hidden-regression diagnostic.}
For method $m$, single-axis attribute $A\in\{\RAC{},\SEX{}\}$, and gap family
$h\in\{\mathrm{DP},\mathrm{EO}\}$, write $I=\INT{}$ and treat lower $G_{A,h}$ as better, with
$G_{A,\mathrm{DP}}=\mathrm{DP}_A$ and $G_{A,\mathrm{EO}}=\mathrm{EO}_A$. We admit a row to the denominator only when a method improves a single-axis gap by at least 1.0
percentage point (\pp{}) relative to ERM, that is
$G_{A,h}(\mathrm{ERM},u,r)-G_{A,h}(m,u,r)\ge 0.01$. Among those rows, we call the method \emph{hidden-regressive} when it worsens the corresponding
intersectional gap or reduces intersectional WGA:
\begin{equation}
  G_{I,h}(m,u,r)-G_{I,h}(\mathrm{ERM},u,r) \ge 0.01,
  \quad\text{or}\quad
  \mathrm{WGA}_{I}(\mathrm{ERM},u,r)-\mathrm{WGA}_{I}(m,u,r) \ge 0.005.
  \label{eq:hidden}
\end{equation}
Informally, we look only at rows where a method has already shown the improvement an audit would
report on race or sex, and we ask whether that same row shows an intersectional cell getting worse.
We chose these reporting thresholds before running the analysis. The 1.0 \pp{} gap threshold avoids
counting tiny DP or EO movements as substantive. We made the 0.5 \pp{} WGA threshold more sensitive because it
measures direct accuracy loss in a worst retained intersectional cell.

\paragraph{The conclusion-flip diagnostic.}
Writing $\mathbf{1}\{\cdot\}$ for the indicator of an event,
$F_c(u,r)=\mathbf{1}\{\widehat{m}_c(u,r) \neq \widehat{m}_c(u,r_0)\}$ is a strict conclusion flip:
missingness changed the selected method relative to its complete-label oracle. The \emph{seed-pair
null}
$F^{\mathrm{seed}}_c(t,q,s,s')=\mathbf{1}\{\widehat{m}_c(t,q,s,r_0) \neq \widehat{m}_c(t,q,s',r_0)\}$
applies the same event definition to two complete-label audits of the same task and split that
differ only in their run seeds $s$ and $s'$. The
matched excess flip rate is
\begin{equation}
  \Delta_c(r)
  =
  \mathbb{E}[F_c(u,r)]
  -
  \mathbb{E}[F^{\mathrm{seed}}_c(t,q,s,s')],
  \label{eq:matched-excess}
\end{equation}
averaged within matched task, split, protected view, candidate set, and criterion cells before
aggregation. Positive $\Delta_c(r)$ means missingness exceeds the complete-label seed floor.
Nonpositive values mean the setting does not move recommendations beyond variation already present
when every label is available.

\paragraph{Asymmetry between the two arms.}
The two arms do not hold the same things fixed. Within a seed, every missingness scenario is evaluated
on identical sampled rows, so the missing-label arm isolates the mask. The seed-null arm does not
hold them fixed, because the run seed also drives context subsampling and per-split row caps, so two complete-label seeds
differ in their sampled data as well as their stochastic audit steps. That baseline therefore
measures the variability of a full audit rerun, where algorithmic randomness is only one term, and a
sample-fixed version would be smaller. We treat this as the main threat to \S\ref{sec:calibration}. The same asymmetry applies to the
complete-label ERM null used in \S\ref{sec:hidden}, where it works in the opposite direction. A
tighter null would raise, not lower, the margin by which threshold optimization exceeds it, but would
also narrow the margin by which non-threshold mitigations fall below it.

\paragraph{Denominators.}
The diagnostics count different objects. Strict flips have denominator $N=22{,}400$, from 560
task/split/seed clusters $\times$ ten label-hiding scenarios $\times$ four criteria. Hidden regression has
$N=31{,}154$ single-axis improvement opportunities across non-ERM methods, two protected axes, and
two gap families. Missing-label intervals resample task/split/seed clusters. The seed-null interval
resamples task/split clusters and keeps within-cluster seed pairs together.

\section{Results}

\begin{figure}[!t]
  \centering
  \includegraphics[width=\linewidth]{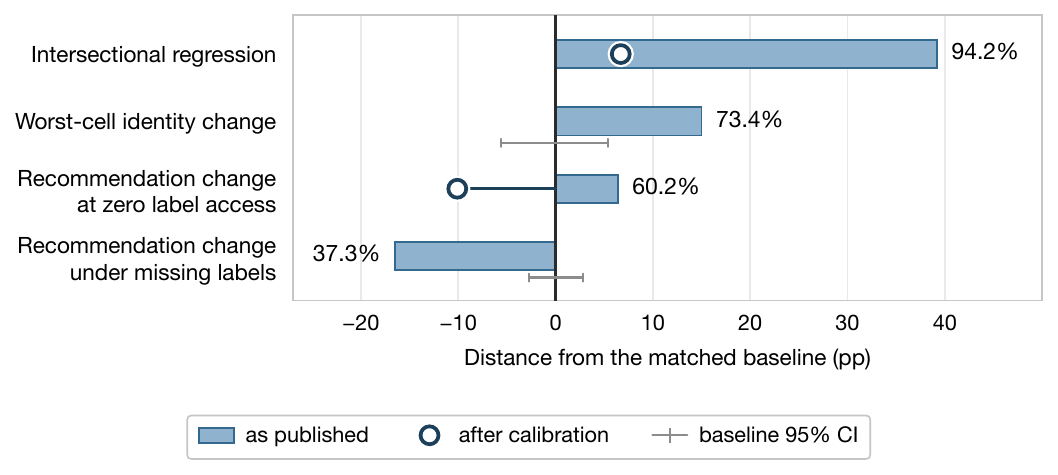}
  \caption{Every rate we publish, shown against the baseline used to interpret it. Zero is that baseline, drawn from the same audit, so a bar extending to the right sits above its baseline and one to the left below it. The top two rows are threshold optimization. Hollow markers give the corrected quantity of \S\ref{sec:hidden} (61.7\%) and \S\ref{sec:calibration} (43.7\%). Three of the four do not stand above their baseline once calibrated. The fourth keeps a 15-point excess.}
  \label{fig:cases}
\end{figure}

Each subsection reports a rate an audit would publish, then the quantity needed to read it, and
\autoref{fig:cases} places all four beside the baselines that read them.

\subsection{Threshold optimization meets its constraint by levelling down}
\label{sec:hidden}

\begin{figure}[!t]
  \centering
  \includegraphics[width=\linewidth]{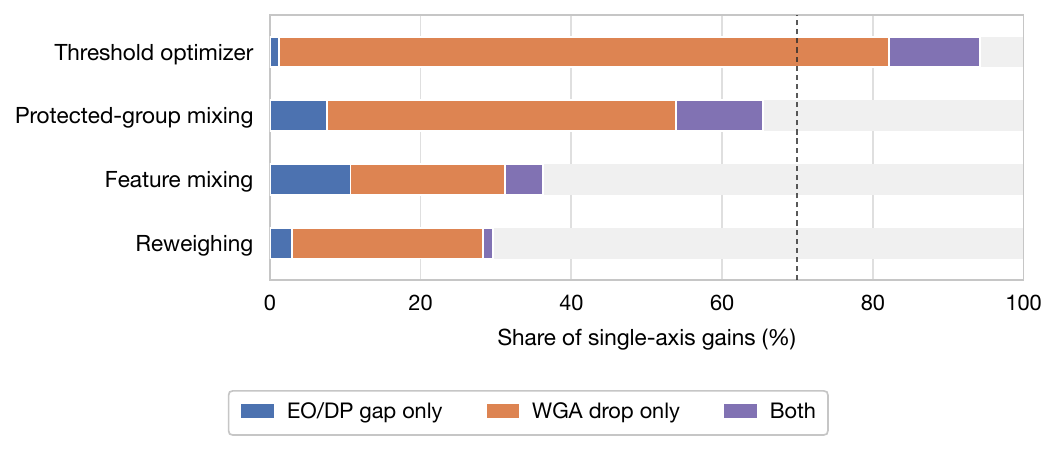}
  \caption{Hidden-regression type by method. Each bar covers that method's single-axis gains, and the filled portion is the share where an intersectional cell regressed. The dashed rule is the 69.9\% rate over all methods. Gains counted are 15{,}147, 6{,}832, 4{,}829, and 4{,}346.}
  \label{fig:hidden}
\end{figure}

\paragraph{Hidden-regression rates.}
We begin with the finding that carries the clearest consequence for people subject to a deployed
model, namely that an audit can report a genuine fairness improvement while the model gets worse for
everyone. Applying the screen in
\autoref{eq:hidden}, a method that improves a race or sex gap by at least 1.0 \pp{} simultaneously
worsens the corresponding intersectional gap or intersectional WGA, a pattern we call \emph{hidden
intersectional regression}, in 69.9\% of those improvement cases (95\% confidence interval, CI: 68.2 to 71.7\%). That aggregate needs calibration. Complete-label
ERM seed pairs already produce a 55.0\% rate under identical thresholds, and excluding threshold
optimization the rate falls to 46.9\%, below that null. Almost all of the excess above the null comes from
a single method. Threshold optimization is hidden-regressive in 94.2\% of its improvement cases,
against 65.4\% for protected-group mixing and 36.2\% and 29.6\% for the two data-level mitigations.
\autoref{fig:hidden} splits each rate by which flag fired, a worsened gap, a worsened worst group, or both.

\paragraph{Concentration of the loss.}
It would be natural to read that 94.2\% as harm moved onto a particular subgroup, but two
measurements show otherwise. Threshold optimization loses 13.3 \pp{} of population accuracy and 13.3
\pp{} of intersectional worst-group accuracy, the same figure. Because plain accuracy rewards
matching the base rate, and the post-processor moves the operating point, we repeat this in balanced
accuracy, the mean of a group's true-positive and true-negative rates, which weights the two classes
equally. There the losses are 5.7 \pp{} for the population and 3.2 \pp{} for the worst-off cell, which
therefore loses \emph{less}.

Recall, however, that worst-group accuracy is a minimum, and that the two minima are frequently taken
over different cells, so neither figure yet says what happened to any particular group. We therefore
ran a paired comparison that tracks each retained cell across the two models and measures the
\emph{concentration} of the loss, meaning how far the hardest-hit cell falls relative to the
population. That excess is a median 7.6 \pp{} in plain accuracy, which looks like concentrated harm.
Nevertheless, it cannot be read directly, since (1) a maximum over roughly a dozen cells of median
support 161 is biased upward, and (2) the statistic inherits the same seed-to-seed movement as every
other rate we report. Two complete-label ERM runs differing only in seed already produce 5.6 \pp{} on
it, and on the invariant metric threshold optimization reaches 5.6 \pp{} against a null of 6.2 \pp{}.
Taking the minimum-based comparison together with the paired one, neither statistic shows
concentration above noise once calibrated, which allows us to conclude that the mechanism here is
levelling down.
\citet{mittelstadt2023unfairness} define levelling down as worsening the better-off without improving
the worse-off, and what we observe is the stronger condition that every cell falls by a similar
amount, and the equalized-odds construction should produce roughly that. The derived predictor of
\citet{hardt2016equality} is confined to the intersection of the per-group
receiver-operating-characteristic polytopes, so an equalized-odds solution cannot sit above the least
favorable group's attainable operating point, and moving down toward that group is how the constraint
gets satisfied. \citet{chen2024multiple} report the same multi-attribute
tension at higher rates on some task--method pairs. With a null, we can say which such rate is anomalous.

\paragraph{Eligibility.}
The 94.2\% does not support any recommendation an auditor would issue, because the
screen applies no eligibility filter and so counts configurations that our own accuracy-constrained
rule rejects. In particular, only 1.7\% of
threshold-optimizer improvement rows fall within 1.0 \pp{} of ERM accuracy, and among those the rate is 61.7\% (95\% CI: 53.5 to 70.1, $n=256$) against the same 55.0\%
baseline, which that interval contains. Among configurations an auditor following our own rule would
consider, we cannot separate threshold optimization from the baseline. The raw 94.2\% is therefore a statement about a
candidate set, and it says far less about any recommendation an audit would actually issue.
Reporting the two figures side by side is the context field 2 of our standard asks for.

\paragraph{Worst-cell identity.}
In 73.4\% of flagged threshold-optimizer cases the worst retained \INT{} cell is not ERM's worst
retained cell (95\% CI: 71.2 to 75.7), against a 58.4\% null (95\% CI: 52.8 to 63.8) from
complete-label runs differing only in seed. Note that the intervals do not overlap, so unlike the
eligibility-filtered rate this excess is established. What survives is worth 15 points against a raw
figure of 73, and the remainder is reshuffling among cells whose accuracies are close enough together
that the ordering between them changes from one seed to the next. The worst-off subgroup is worth reporting, and
so is the rate at which its identity shifts across seeds. Appendices~\ref{app:hidden} and~\ref{app:operating-point} give
failure types, per-axis breakdowns, and the full operating-point recomputation.

\subsection{Missingness rarely moves recommendations beyond seed noise}
\label{sec:calibration}

\paragraph{Seed calibration.}
We turn next to the quantity most audits would report first. We show that most of the missingness
signal a raw sensitivity analysis produces cannot be distinguished from noise the audit already
contains. Strict method choices change in 37.3\% of
missing-label comparisons (95\% CI: 36.0 to 38.6\%), but complete-label seed pairs already change in
53.8\% of unordered comparisons (95\% CI: 51.1 to 56.6\%). Because that baseline also varies the
sampled rows, it is not a floor on algorithmic noise alone and the two arms cannot be equated. What
follows is the weaker claim that hiding labels moves recommendations less than rerunning the whole
audit does. Appendix~\ref{app:diagnostics} reports both diagnostics against their matched baselines
(\autoref{tab:evidence}) and plots the full availability curve against that seed floor
(\autoref{fig:flip-rate}). Flips are most frequent with no observed labels (60.2\%) and fall to
20.1\% at 80\% MCAR availability. After cluster-matching by task, split, and criterion, every MCAR
setting retaining some labels and every matched MNAR setting stays below its seed floor. The single
exception is the 0\% setting (+6.3 \pp{}, 95\% CI: +2.4 to +10.4). Of every number in this analysis
that one is the most easily misread as evidence that missing labels destabilize audits, and it turns
out to be an artifact of candidate equivalence, so we take it apart next.

\paragraph{The no-label endpoint.}
The 0\% endpoint is the one setting exceeding its seed floor, and we show that it should not be read
as a missingness effect, because it responds to the composition of the candidate set. With no
protected labels, reweighing uses all-one weights and protected-group mixing has no observed groups to mix, so
both reproduce ERM metrics \emph{exactly} in all 560 clusters. The selection among them is then decided by
the alphabetical tie breaker, with no measured difference behind it, and the raw rate counts it as a
flip. In 42.0\% of no-label decisions the winner is an exact ERM/reweighing/protected-group-mixing
tie, always returning ERM. Collapsing those candidates into one class drops the rate from 60.2\% to 43.7\%, below
the 53.8\% floor. Overall, we obtain no setting in this benchmark that exceeds complete-label
variation once identical models are counted once (\autoref{fig:flip-rate}).
Candidate-set composition matters for the same reason. Recomputing selection over
\{ERM, reweighing, threshold optimization\} yields a 33.8\% raw flip rate against a 39.8\% restricted
seed null, and 30.6\% among settings retaining some labels. Two margin checks confirm that tie
movement does not account for all of the remaining flips, since 18.1\% survive a one-\pp{} criterion
margin and 13.2\% a two-\pp{} margin (Appendix~\ref{app:diagnostics}). Taking the equivalence check
together with the restricted-candidate recomputation, we conclude that the no-label endpoint reports
on the candidate set an auditor assembled and not on the labels they lost.

\subsection{What the tradeoffs and a second base learner do not change}
\label{sec:tradeoffs}
\label{sec:rf}

\paragraph{Two bounding checks.}
We run two checks to bound how far the preceding results generalize. The first asks what an auditor
gives up by picking any candidate at all, and finds that none dominates. Feature mixing is alone in
having a positive average accuracy delta, at +0.16 \pp{}, and it also carries the highest unsafe-flag
rate at 30.7\%. Reweighing is close to neutral on both counts. Threshold optimization reduces the
intersectional EO gap by 8.7 \pp{} at a cost of 13.3 \pp{} of accuracy, or 5.7 \pp{} on
the balanced-accuracy scale its own objective optimizes. The second check swaps the base learner for
a random forest, holding everything else fixed. Recommendation churn is directionally higher, 41.1\%
against 37.3\% overall, while the subgroup result barely moves, 69.4\% against 69.9\% overall and
93.5\% against 94.2\% for threshold optimization, so the finding that carries \S\ref{sec:hidden} does
not depend on the learner that produced it. Appendices~\ref{app:rf} and~\ref{app:heterogeneity}
report both panels in full.

\section{The \ADS{}}
\label{sec:standard}

\paragraph{From cases to fields.}
Every result above took the same form. In each, a published rate pointed one way, and a second
quantity that the audit had already computed pointed the other. We list the cases in
\autoref{tab:cases} (Appendix~\ref{app:diagnostics}), each with the field it implicates. None
required extra data collection, and each turned on a field the report omitted.
\autoref{tab:standard} states them as nine disclosures, grouped by the question each answers, each
with the misreading it prevents. The grouping is the order an oversight reader
should use. Fields 1--3 establish what the audit compared, so a changed recommendation can be
attributed to something. Fields 4--6 establish what it should be compared
against, so a rate can be judged against the noise floor and the label-access mechanism that
produced it. Fields 7--9 establish who bears the cost, the only part of the report that speaks directly to the
people the system acts on.

\begin{table}[!ht]
  \captionsetup{skip=7pt}
  \caption{The \ADS{}: the nine fields an audit report must contain so its published rates can be read correctly, grouped by the question each answers. The benchmark cases that motivate the standard appear in \autoref{tab:cases} (Appendix~\ref{app:diagnostics}).}
  \label{tab:standard}
  \centering
  \footnotesize
  \setlength{\tabcolsep}{5pt}
  \renewcommand{\arraystretch}{1.02}
  \begin{tabularx}{\linewidth}{@{}>{\raggedright\arraybackslash}p{0.185\linewidth}>{\raggedright\arraybackslash}X>{\raggedright\arraybackslash}X@{}}
    \toprule
    \textbf{Field} & \textbf{What the report should state} & \textbf{Misreading it prevents}\\
    \midrule
    \adspanelfirst{A.}{What did the audit compare?}
    \adsrow{1}{Candidate set}{Every mitigation the selector could return, near-equivalents included}{A candidate-set change read as a real effect}
    \adsrow{2}{Criterion, margin}{Selection rule, eligibility filter, accuracy tolerance, tie breaker}{A recommendation read without its rule}
    \adsrow{3}{Equivalence class}{Candidates with identical predictions under the available labels}{A renamed identical model read as a real change}
    \adspanel{B.}{What is that comparison worth?}
    \adsrow{4}{Seed floor}{The same rate recomputed across seeds, all labels present}{Seed variation read as a missing-label effect}
    \adsrow{5}{Realized availability}{Observed availability per scenario, and unknown-record counts}{An availability claim taken on trust}
    \adsrow{6}{Missingness mechanism}{Whether loss is random or group-dependent, and who loses labels}{A group-dependent loss read as uniform}
    \adspanel{C.}{Who bears the cost?}
    \adsrow{7}{Intersectional gaps}{DP and EO gaps on intersections, beside every single-axis gap}{A regression hidden behind a single-axis gain}
    \adsrow{8}{Worst-group accuracy}{Minimum per-group accuracy on retained cells, beside the population}{Levelling down read as targeted harm}
    \adsrow{9}{Worst-cell identity}{Which subgroup is worst off, and how often it changes}{A shifting worst-off group read as stable}
    \bottomrule
  \end{tabularx}
\end{table}

\paragraph{Relation to existing disclosure requirements.}
Most of these fields are already familiar. Model cards and datasheets established the convention that a system is documented together with
the conditions under which it was evaluated \citep{mitchell2019model,gebru2021datasheets}. The
audit-governance literature has since argued that the binding constraint is the absence of agreed
standards for what an audit must report, more than the availability of audit methods
\citep{costanzachock2022audits}. New York City's Local Law 144 goes further. A published
bias audit there must report impact ratios by sex, by race and ethnicity, and by intersectional
category, together with the count of individuals whose demographic category is unknown and the
provenance of the audit data. Its implementing rules, not the statute, permit excluding categories
below 2\% of the data. That is exactly the kind of retention rule that field 8 asks auditors to state
\citep{nyc2021ll144,nycdcwp2023rules}.
Fields 5, 7, and 8 therefore largely restate obligations that already bind regulated deployers in at
least one jurisdiction, and our contribution there is evidence for why they matter. Three fields are not covered by any standard we are aware of. They are the equivalence-class check (3), which distinguishes a
changed method name from changed predictions, the complete-label seed floor (4), without which no
disclosed rate can be judged, and worst-cell identity (9), which separates a uniform cost from harm
transferred onto a different community.

Procurement reviewers and regulators work from the audit report, not from the code that produced it,
so oversight can check only the report itself. We release our code, configs, and the tables
behind every number with the paper.

\paragraph{Cost of compliance.}
The standard is cheap to satisfy. Fields 1--3 are properties of the audit specification,
fields 4--6 need one rerun under a different seed plus a record of the masking procedure, and fields
7--9 need existing metrics evaluated on intersectional cells as well as on the reported axes. None
requires demographic data an institution does not already hold, so only convention stands in the way
of reporting them. Reporting fields 7--9 would have caught the
threshold-optimizer behavior of \S\ref{sec:hidden} from the audit's own outputs.

\section{Conclusion}

Fairness audits are a key component of responsible machine-learning deployment. Yet what a published
audit rate means, when the protected labels behind it are incomplete, is still poorly understood. In
this work, we focused on the rates such an audit reports and on the baselines an oversight reader
needs to read them. Across four ACS/Folktables tasks we found that five of the rates we report
change meaning once we pair each with a baseline drawn from the same audit. As a
result, protected-label missingness, the condition institutions most often apologize for, matters less
than publishing a rate without the quantity that interprets it. We also showed that the
threshold optimizer's subgroup harm is not concentrated on one cell, since the loss falls on the
population as heavily and disappears into its baseline among the configurations an auditor would
accept. Post-processing that enforces a group constraint is therefore not interchangeable with
data-level mitigation, and fields 7--9 should be mandatory wherever it is used. Overall, our results highlight that a rate an audit publishes has no interpretation on its own,
and that the quantity supplying one is already available to the auditor.

\paragraph{Limitations and future work.}
For the reason given in \S\ref{sec:setup}, our seed-pair baseline measures the variability of a full
audit rerun, in which algorithmic noise is only one part, so a sample-fixed rerun could place some
positive-availability scenarios above it.
Missingness is also simulated through a single majority-versus-rest MNAR mechanism. We have
validated the standard against the misreadings we can observe in this benchmark, but we have not yet
tested it with the institutional readers it is meant to help.

\clearpage
\bibliographystyle{plainnat}
\bibliography{references}

\clearpage
\appendix
\captionsetup[table]{font=small, labelfont=bf, skip=6pt}
\setlength{\floatsep}{18pt plus 4pt minus 2pt}
\setlength{\textfloatsep}{20pt plus 4pt minus 2pt}
\setlength{\intextsep}{18pt plus 4pt minus 2pt}

\section{Audit configuration}
\label{app:config}

\paragraph{Row caps and mixing.}
This appendix records the configuration behind \S\ref{sec:setup}. Row caps were chosen to keep
contexts comparable and memory bounded for one-hot encoded logistic and random-forest runs. Feature
mixing and protected-group mixing use $\lambda\sim\mathrm{Beta}(0.4,0.4)$ and a 1.0 mix ratio, so
synthetic rows stay near observed examples while adding about one row per observed row. Reweighing
uses Kamiran--Calders group-label weights \citep{kamiran2012data}. Threshold optimization uses the
group-specific decision thresholds that satisfy an equalized-odds constraint while maximizing
balanced accuracy, fitted on a 25\% validation split held out from training. Groups whose validation
rows carry only one label value admit no threshold and are merged deterministically into the largest
well-conditioned group before fitting. Its objective is balanced accuracy, while the other candidates
maximize plain accuracy. Appendix~\ref{app:operating-point} works out what this difference means for
the worst-cell comparison.

\paragraph{Seeds and base learners.}
Primary logistic-regression results use five run seeds. The base learner's own initialization is
held fixed across all five, while the run seed controls row subsampling, missingness masks, mixing
draws, and the threshold optimizer's validation split, so seed-to-seed movement reflects the
stochastic audit steps, with the estimator itself re-initialized identically every time. The random-forest validation
run uses 120 trees, maximum depth 18, a minimum of 8 training rows per leaf, and the same five
seeds. The 560 task/split/seed clusters comprise four tasks crossed with temporal and
leave-one-state-out geographic splits and five seeds. The eleven missingness scenarios are the
complete-label case plus seven MCAR and three matched MNAR variants.

\FloatBarrier
\section{Conclusion-flip calibration}
\label{app:diagnostics}

\paragraph{What the flip rate needs alongside it.}
The strict conclusion-flip rate is the first quantity we examine, because its headline percentage
can only be interpreted alongside its denominator and complete-label seed baseline.
\autoref{tab:diagnostics} collects the main rates the paper uses. After reporting
those quantities, we check margin-qualified flips (\autoref{tab:margin-diagnostics}), no-label
equivalence, restricted candidate sets, and per-criterion rates. Taken together, these diagnostics
distinguish flips tied to missing protected labels from recommendation changes that already appear
under complete labels.

\begin{table}[!htbp]
  \caption{Five rates from this benchmark, as an audit would first compute them and after the calibration each requires. The final column gives the field of the \ADS{} that supplies the missing context.}
  \label{tab:cases}
  \centering
  \small
  \setlength{\tabcolsep}{5pt}
  \renewcommand{\arraystretch}{1.12}
  \begin{tabular}{@{}p{0.30\linewidth}rp{0.40\linewidth}c@{}}
    \toprule
    Quantity & As computed & After calibration & Field \\
    \midrule
    Recommendation changes under missing labels & 37.3\% & below a 53.8\% same-audit rerun baseline & 4 \\
    The same, at zero label access & 60.2\% & 43.7\% counting identical candidates once & 3 \\
    Intersectional regression, threshold optimization & 94.2\% & 61.7\% among eligible configurations, an interval covering the baseline & 2 \\
    Largest same-cell accuracy drop, above the population & +7.6 \pp{} & +5.6 \pp{} baseline; below it on the invariant metric & 8 \\
    Change in which subgroup is worst off & 73.4\% & 58.4\% baseline, a 15-point excess that does survive & 9 \\
    \bottomrule
  \end{tabular}
\end{table}
\appendixtablegap

\begin{table}[!htbp]
  \caption{Audit evidence at a glance. Panel A counts audit-selection decisions and reports the matched excess over the complete-label seed-pair null; panel B counts single-axis improvement cases and decomposes them into intersectional gap worsening, WGA drops, and both. Panel A shows four of the ten missingness scenarios; \autoref{tab:criterion-flips} gives all of them. The no-label row is the only positive excess in panel A and is marked because it is not a like-for-like missingness comparison; see \S\ref{sec:calibration} and \autoref{tab:no-label-equivalence}.}
  \label{tab:evidence}
  \centering
  \small
  \begingroup
\setlength{\tabcolsep}{4.1pt}
\renewcommand{\arraystretch}{1.04}
\begin{tabular}{@{}lrrrrrr@{}}
\toprule
Check & Rate & $\Delta$ seed & Gap & WGA & Both & n \\
 & (\%) & (pp) & (\%) & (\%) & (\%) & \\
\midrule
\multicolumn{7}{@{}l}{\textit{A. Method-selection stability}} \\
Seed-pair null & 53.8 & n/a & n/a & n/a & n/a & 4480 \\
No-label MCAR$^{\dagger}$ & 60.2 & \textbf{+6.3} & n/a & n/a & n/a & 448 \\
MCAR 10\% & 42.2 & -11.7 & n/a & n/a & n/a & 448 \\
MCAR 80\% & 20.1 & -33.7 & n/a & n/a & n/a & 448 \\
MNAR 10\% & 44.6 & -9.2 & n/a & n/a & n/a & 448 \\
\addlinespace[3pt]
\multicolumn{7}{@{}l}{\textit{B. Single-axis gains}} \\
Overall & \textbf{69.9} & n/a & 13.7 & 65.6 & 9.4 & 31154 \\
Excl. threshold opt. & 46.9 & n/a & 14.0 & 39.6 & 6.8 & 16007 \\
Threshold optimizer & \textbf{94.2} & n/a & 13.4 & \textbf{93.0} & 12.1 & 15147 \\
Protected mixing & 65.4 & n/a & 19.1 & 57.9 & 11.5 & 6832 \\
\bottomrule
\addlinespace[2pt]
\multicolumn{7}{@{}p{0.60\linewidth}@{}}{\footnotesize $^{\dagger}$Not a like-for-like missingness comparison. With no observed labels, reweighing and protected-group mixing reproduce ERM exactly, so this rate is inflated by the alphabetical tie breaker renaming an identical model; collapsing ERM-equivalent candidates leaves it below the seed null.} \\
\end{tabular}
\endgroup

\end{table}
\appendixtablegap

\begin{table}[!htbp]
  \caption{Main diagnostic rates used throughout the paper. Rates and confidence intervals are percentages; confidence intervals use a cluster bootstrap over task/split/seed clusters. Conclusion-flip rows count audit-selection decisions, while hidden-regression rows count single-axis improvement cases.}
  \label{tab:diagnostics}
  \centering
  \small
  \setlength{\tabcolsep}{4pt}
  \renewcommand{\arraystretch}{1.16}
  \begin{tabular}{lllll}
\toprule
Diagnostic & Rate (\%) & 95\% CI low (\%) & 95\% CI high (\%) & N \\
\midrule
Strict conclusion flips overall & \textbf{37.3} & 36.0 & 38.6 & 22400 \\
Margin-filtered flips (both audits >=1 pp) & \textbf{18.1} &  &  & 22400 \\
Full-label seed-pair null flips & \textbf{53.8} &  &  & 4480 \\
Strict conclusion flips at 80\% availability & \textbf{20.1} & 18.3 & 22.0 & 2240 \\
Strict conclusion flips at 50\% availability & \textbf{33.0} & 31.3 & 34.7 & 4480 \\
Strict conclusion flips at 20\% availability & \textbf{40.7} & 39.0 & 42.5 & 4480 \\
Strict conclusion flips at 10\% availability & \textbf{43.4} & 41.6 & 45.1 & 4480 \\
Strict conclusion flips at 0\% availability & \textbf{60.2} & 57.8 & 62.5 & 2240 \\
Hidden intersectional regressions overall & \textbf{69.9} & 68.2 & 71.7 & 31154 \\
Hidden regressions excluding threshold optimizer & \textbf{46.9} & 44.2 & 49.4 & 16007 \\
ERM seed-pair hidden-regression null & \textbf{55.0} &  &  & 2714 \\
\bottomrule
\end{tabular}

\end{table}
\appendixpagegap

\begin{figure}[!htbp]
  \centering
  \includegraphics[width=0.9\linewidth]{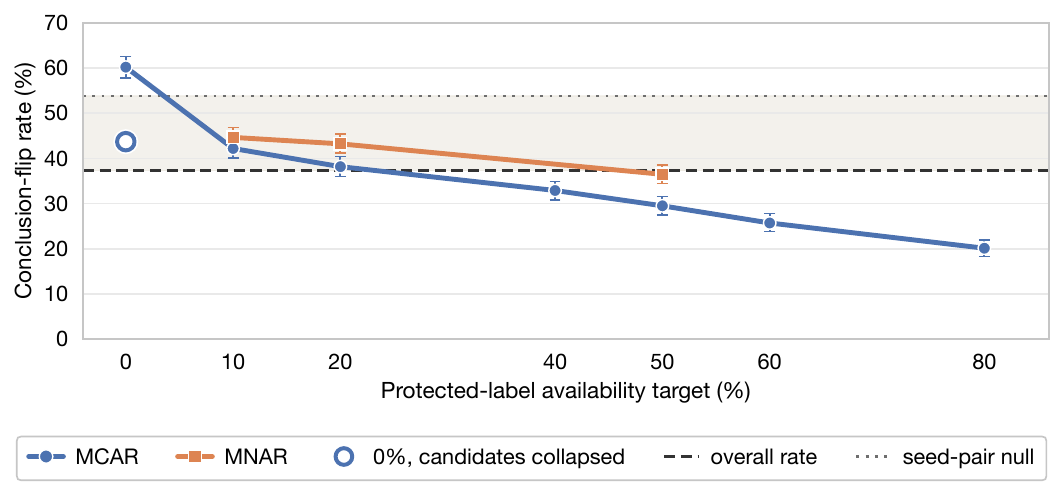}
  \caption{Strict conclusion-flip rates with cluster-bootstrap 95\% intervals. Complete-label audits serve as the oracle reference. The dashed line is the overall strict flip rate (37.3\%), the dotted line the full-label seed-pair null (53.8\%), and the band between them the zone where a flip rate is indistinguishable from seed noise. Availability is the nominal target for both MCAR and MNAR. The 0\% endpoint is not a like-for-like missingness comparison: reweighing and protected-group mixing reproduce ERM exactly there, so the filled marker (60.2\%) largely counts the alphabetical tie breaker renaming an identical model. The hollow marker (43.7\%) recomputes the same rate with ERM-equivalent candidates collapsed into one class, which places it below the seed-pair null.}
  \label{fig:flip-rate}
\end{figure}
\appendixtablegap

\noindent The cluster-matched comparison behind \S\ref{sec:calibration} appears in
\autoref{tab:seed-matched-flips}, and \autoref{tab:seed-null} reports the complete-label seed
baseline it is matched against.

\begin{table}[!htbp]
  \caption{Cluster-matched comparison of missing-label strict conclusion flips against the complete-label seed-pair null. Matching is by task, split, and selection criterion; the seed-null column is the same global complete-label null in every row, and positive deltas indicate missingness exceeds the complete-label seed floor.}
  \label{tab:seed-matched-flips}
  \centering
  \small
  \begingroup
\setlength{\tabcolsep}{5pt}
\renewcommand{\arraystretch}{1.12}
\begin{tabular}{@{}lrrrrlr@{}}
\toprule
Mechanism & Availability & Missing & Seed null & $\Delta$ & 95\% CI $\Delta$ & n \\
 & (\%) & flip (\%) & flip (\%) & (pp) & (pp) & cells \\
\midrule
MCAR & 0 & 60.2 & 53.8 & \textbf{+6.3} & [+2.4, +10.4] & 448 \\
MCAR & 10 & 42.2 & 53.8 & -11.7 & [-14.4, -9.0] & 448 \\
MCAR & 20 & 38.2 & 53.8 & -15.7 & [-18.5, -12.9] & 448 \\
MCAR & 40 & 32.9 & 53.8 & -20.9 & [-23.5, -18.4] & 448 \\
MCAR & 50 & 29.5 & 53.8 & -24.3 & [-26.9, -21.4] & 448 \\
MCAR & 60 & 25.7 & 53.8 & -28.1 & [-30.7, -25.2] & 448 \\
MCAR & 80 & 20.1 & 53.8 & -33.7 & [-36.3, -30.9] & 448 \\
MNAR & 10 & 44.6 & 53.8 & -9.2 & [-12.1, -6.2] & 448 \\
MNAR & 20 & 43.2 & 53.8 & -10.6 & [-13.2, -7.9] & 448 \\
MNAR & 50 & 36.5 & 53.8 & -17.4 & [-20.0, -14.7] & 448 \\
\bottomrule
\end{tabular}
\endgroup

\end{table}
\appendixtablegap

\begin{table}[!htbp]
  \caption{Complete-label seed-pair null for strict method selection. It quantifies how often a matched audit changes recommendations across seeds before any protected labels are hidden.}
  \label{tab:seed-null}
  \centering
  \small
  \renewcommand{\arraystretch}{1.16}
  \begin{tabular}{lll}
\toprule
Criterion & Rate (\%) & N \\
\midrule
\textbf{Overall} & 53.8 & 4480 \\
Accuracy-constrained EO & \textbf{66.7} & 1120 \\
Demographic parity & 33.6 & 1120 \\
Worst-group accuracy & 66.1 & 1120 \\
Intersectional EO & 49.0 & 1120 \\
\bottomrule
\end{tabular}

\end{table}
\appendixtablegap

\begin{table}[!htbp]
  \caption{Margin diagnostics for strict conclusion flips. Margin filters remove cases in which either audit barely prefers its selected method; the one-\pp{} row is the main robustness check.}
  \label{tab:margin-diagnostics}
  \centering
  \small
  \renewcommand{\arraystretch}{1.16}
  \begin{tabular}{llll}
\toprule
Margin filter & Rate (\%) & N & n\_flips \\
\midrule
\textbf{strict arg-max} & \textbf{37.3} & 22400 & \textbf{8358} \\
both >= 0.5 pp & 21.7 & 22400 & 4867 \\
\textbf{both >= 1 pp} & \textbf{18.1} & 22400 & \textbf{4065} \\
both >= 2 pp & 13.2 & 22400 & 2956 \\
\bottomrule
\end{tabular}

\end{table}
\appendixpagegap

The 0\% protected-label setting needs a separate check because several candidate methods become
empirically indistinguishable from ERM. \autoref{tab:no-label-equivalence} and
\autoref{tab:restricted-candidates} separate this candidate-set collapse from broader missingness
sweeps, supporting fields 1 and 3 of the \ADS{}.

\begin{table}[!htbp]
  \caption{No-label MCAR equivalence check. At 0\% protected-label availability, ERM, reweighing, and protected-group mixing have numerically identical metrics in all 560 task/split/seed clusters, verified to machine precision across every audit metric. Collapsing ERM-equivalent methods into one class separates prediction changes from exact ties resolved only by the alphabetical tie breaker. The overall strict rate falls from 60.2\% to 43.7\%, below the 53.8\% complete-label seed-pair null.}
  \label{tab:no-label-equivalence}
  \centering
  \small
  \begingroup
\setlength{\tabcolsep}{4.2pt}
\renewcommand{\arraystretch}{1.12}
\begin{tabular}{@{}lrrrr@{}}
\toprule
Criterion & N & Strict flip (\%) & ERM-equivalence flip (\%) & Exact-tie selected by name (\%) \\
\midrule
Accuracy-constrained EO & 560 & 58.9 & 29.8 & 45.9 \\
Demographic parity & 560 & 75.2 & 64.3 & 44.5 \\
Worst-group accuracy & 560 & 46.2 & 30.5 & 37.0 \\
Intersectional EO & 560 & 60.4 & 50.2 & 40.5 \\
\midrule
\textbf{Overall} & \textbf{2240} & \textbf{60.2} & \textbf{43.7} & \textbf{42.0} \\
\bottomrule
\end{tabular}
\endgroup

\end{table}
\appendixpagegap

\begin{table}[!htbp]
  \caption{Restricted candidate-set conclusion flips. Selection is recomputed from the existing metric rows using only ERM, reweighing, and threshold optimization. Restricting the set separates flips from stable methods and flips created by bespoke candidates that collapse to ERM under no-label access.}
  \label{tab:restricted-candidates}
  \centering
  \footnotesize
  \resizebox{\linewidth}{!}{\begingroup
\setlength{\tabcolsep}{4.5pt}
\renewcommand{\arraystretch}{1.1}
\begin{tabular}{@{}lllrrlr@{}}
\toprule
Subset & Mechanism & Availability & Restricted & Seed null & $\Delta$ & n \\
 & & & flip (\%) & flip (\%) & (pp) & \\
\midrule
All label-hiding scenarios & ALL & all & 33.8 & 39.8 & -5.9 & 22400 \\
Positive availability only & ALL & >0 & 30.6 & 39.8 & -9.2 & 20160 \\
By setting & MCAR & 0\% & 62.9 & 39.8 & \textbf{+23.2} & 448 \\
By setting & MCAR & 10\% & 37.7 & 39.8 & -2.0 & 448 \\
By setting & MCAR & 20\% & 35.0 & 39.8 & -4.8 & 448 \\
By setting & MCAR & 40\% & 28.6 & 39.8 & -11.1 & 448 \\
By setting & MCAR & 50\% & 25.0 & 39.8 & -14.7 & 448 \\
By setting & MCAR & 60\% & 20.8 & 39.8 & -19.0 & 448 \\
By setting & MCAR & 80\% & 15.4 & 39.8 & -24.4 & 448 \\
By setting & MNAR & 10\% & 41.4 & 39.8 & \textbf{+1.7} & 448 \\
By setting & MNAR & 20\% & 39.1 & 39.8 & -0.6 & 448 \\
By setting & MNAR & 50\% & 32.1 & 39.8 & -7.6 & 448 \\
\bottomrule
\end{tabular}
\endgroup
}
\end{table}
\appendixpagegap

\begin{table}[!htbp]
  \caption{Per-criterion strict conclusion-flip rates by missingness mechanism and protected-label availability. Criterion breakdowns identify which selection rules account for aggregate availability curves.}
  \label{tab:criterion-flips}
  \centering
  \small
  \begingroup
\setlength{\tabcolsep}{4.6pt}
\renewcommand{\arraystretch}{1.12}
\begin{tabular}{@{}lrrrrrrrr@{}}
\toprule
\multicolumn{9}{@{}l}{\textit{MCAR}} \\
Criterion & 0\% & 10\% & 20\% & 40\% & 50\% & 60\% & 80\% & n \\
\midrule
Acc.-constrained EO & 58.9 & 47.3 & 42.1 & 34.5 & 33.2 & 28.0 & 20.0 & 560 \\
Demographic parity & \textbf{75.2} & 37.0 & 31.8 & 28.0 & 24.3 & 20.2 & 17.3 & 560 \\
Intersectional EO & 60.4 & 43.6 & 41.2 & 34.5 & 33.9 & 29.5 & 20.9 & 560 \\
Worst-group accuracy & 46.2 & 40.9 & 37.5 & 34.6 & 26.6 & 25.2 & 22.3 & 560 \\
\bottomrule
\end{tabular}
\vspace{1.35em}
\begin{tabular}{@{}lrrrr@{}}
\toprule
\multicolumn{5}{@{}l}{\textit{MNAR}} \\
Criterion & 10\% & 20\% & 50\% & n \\
\midrule
Acc.-constrained EO & 48.8 & 47.0 & 43.8 & 560 \\
Demographic parity & 41.1 & 40.0 & 28.9 & 560 \\
Intersectional EO & 44.5 & 43.6 & 38.2 & 560 \\
Worst-group accuracy & 44.3 & 42.3 & 35.0 & 560 \\
\bottomrule
\end{tabular}
\endgroup

\end{table}
\appendixpagegap

\FloatBarrier
\section{Hidden intersectional-regression diagnostics}
\label{app:hidden}

\paragraph{Decomposing the result.}
Hidden regression has several components, so this appendix decomposes the result from several angles.
We separate failures by type (\autoref{tab:hidden-decomp}), compare them with an ERM seed-pair null
(\autoref{tab:hidden-null}), expand rates across metric families (\autoref{tab:hidden-by-method}),
measure WGA-drop magnitudes (\autoref{tab:wga-magnitudes}), and check whether the worst intersectional
cell changes (\autoref{tab:wga-identity}). All rows follow the screening rule defined in
\autoref{eq:hidden}. These tables are the evidence base for fields 7--9 of the \ADS{}.

\appendixtablegap

\begin{table}[!htbp]
  \caption{Decomposition of hidden intersectional regression flags. Gap denotes intersectional demographic-parity or equalized-odds worsening; WGA denotes an intersectional worst-group-accuracy drop. WGA-only failures dominate threshold optimization.}
  \label{tab:hidden-decomp}
  \centering
  \small
  \begingroup
\setlength{\tabcolsep}{3.8pt}
\renewcommand{\arraystretch}{1.12}
\begin{tabular}{@{}lrrrrrrr@{}}
\toprule
Method & Hidden & Gap & WGA drop & Gap only & WGA only & Both & n \\
\midrule
\textbf{Overall} & 69.9 & 13.7 & 65.6 & 4.3 & 56.2 & 9.4 & 31154 \\
Feature mixing & 36.2 & 15.7 & 25.5 & \textbf{10.7} & 20.5 & 5.0 & 4829 \\
Reweighing & 29.6 & 4.3 & 26.7 & 2.9 & 25.3 & 1.4 & 4346 \\
Protected-group mixing & 65.4 & \textbf{19.1} & 57.9 & 7.6 & 46.4 & 11.5 & 6832 \\
\textbf{Threshold optimizer} & \textbf{94.2} & 13.4 & \textbf{93.0} & 1.3 & \textbf{80.9} & \textbf{12.1} & 15147 \\
\bottomrule
\end{tabular}
\endgroup

\end{table}
\begin{table}[!htbp]
  \caption{ERM seed-pair null for hidden intersectional regression flags. It applies the same single-axis-improvement and intersectional-regression thresholds to ordered complete-label ERM seed pairs.}
  \label{tab:hidden-null}
  \centering
  \small
  \begingroup
\setlength{\tabcolsep}{4pt}
\renewcommand{\arraystretch}{1.1}
\begin{tabular}{@{}lrrrrrrr@{}}
\toprule
Comparison & Hidden & Gap & WGA drop & Gap only & WGA only & Both & n \\
\midrule
\textbf{ERM seed-pair null} & \textbf{55.0} & \textbf{26.8} & \textbf{41.7} & \textbf{13.2} & \textbf{28.2} & \textbf{13.6} & 2714 \\
\bottomrule
\end{tabular}
\endgroup

\end{table}
\appendixpagegap

\begin{table}[!htbp]
  \caption{Hidden-regression rates by method, single-axis attribute, and metric family. DP denotes demographic parity, EO denotes equalized odds, and $n$ counts single-axis improvement rows. Bold marks the largest hidden-regression rate within each method.}
  \label{tab:hidden-by-method}
  \centering
  \small
  \setlength{\tabcolsep}{4pt}
  \renewcommand{\arraystretch}{1.14}
  \begin{tabular}{lllll}
\toprule
Method & Single axis & Metric family & Hidden rate (\%) & n \\
\midrule
Feature mixing & RAC1P & DP & 32.3 & 2112 \\
Feature mixing & RAC1P & EO & 35.2 & 2310 \\
Feature mixing & SEX & DP & 50.0 & 110 \\
Feature mixing & SEX & EO & \textbf{66.7} & 297 \\
Protected-group mixing & RAC1P & DP & 61.1 & 2439 \\
Protected-group mixing & RAC1P & EO & 65.3 & 2556 \\
Protected-group mixing & SEX & DP & \textbf{72.7} & 1002 \\
Protected-group mixing & SEX & EO & 69.8 & 835 \\
Reweighing & RAC1P & DP & 27.8 & 1351 \\
Reweighing & RAC1P & EO & 27.2 & 1335 \\
Reweighing & SEX & DP & 31.1 & 1076 \\
Reweighing & SEX & EO & \textbf{36.3} & 584 \\
\textbf{Threshold optimizer} & RAC1P & DP & 94.1 & 4305 \\
\textbf{Threshold optimizer} & RAC1P & EO & 93.4 & 4232 \\
\textbf{Threshold optimizer} & SEX & DP & \textbf{95.4} & 3009 \\
\textbf{Threshold optimizer} & SEX & EO & 94.4 & 3601 \\
\bottomrule
\end{tabular}

\end{table}
\appendixpagegap

Because WGA-only drops account for most of the hidden-regression signal, the next two tables focus on
them. \autoref{tab:wga-magnitudes} reports the size of those losses, while \autoref{tab:wga-identity}
checks whether the worst intersectional cell itself changes.

\begin{table}[!htbp]
  \caption{Magnitude of WGA drops among hidden-regression rows where the WGA flag fires. Tail statistics distinguish small intersectional shifts from the large WGA losses concentrated in threshold optimization.}
  \label{tab:wga-magnitudes}
  \centering
  \small
  \begingroup
\setlength{\tabcolsep}{4.2pt}
\renewcommand{\arraystretch}{1.12}
\begin{tabular}{@{}lrrrrrrr@{}}
\toprule
Comparison & n & Median & P75 & P90 & $\geq$1 pp & $\geq$2 pp & $\geq$5 pp \\
\midrule
\textbf{Overall} & 20428 & 7.98 & 16.44 & 28.02 & 92.6 & 81.6 & 63.3 \\
Excluding threshold optimizer & 6345 & 1.95 & 3.41 & 6.25 & 78.0 & 48.2 & 15.2 \\
Feature mixing & 1232 & 1.37 & 2.15 & 2.85 & 70.5 & 29.5 & 8.0 \\
Reweighing & 1159 & 1.36 & 3.03 & 5.94 & 63.0 & 41.5 & 15.2 \\
Protected-group mixing & 3954 & 2.33 & 3.95 & 6.65 & 84.7 & 56.1 & 17.4 \\
\textbf{Threshold optimizer} & 14083 & \textbf{12.34} & \textbf{20.73} & \textbf{33.34} & \textbf{99.1} & \textbf{96.6} & \textbf{85.0} \\
\bottomrule
\end{tabular}
\endgroup

\end{table}
\appendixtablegap

\begin{table}[!htbp]
  \caption{Worst-cell identity among hidden-regression rows where the WGA flag fires. Differs means a method's worst retained \INT{} cell is not the ERM worst retained \INT{} cell in a matched task/split/seed/missingness row; median $n$ is the test-row support of that worst cell. For threshold optimization the rate is 73.4\% (95\% CI: 71.2 to 75.7) against a complete-label ERM seed-pair null of 58.4\% (95\% CI: 52.8 to 63.8); see \autoref{tab:intervals}.}
  \label{tab:wga-identity}
  \centering
  \small
  \begingroup
\setlength{\tabcolsep}{4.4pt}
\renewcommand{\arraystretch}{1.12}
\begin{tabular}{@{}lrrrrrr@{}}
\toprule
Comparison & n & Differs & Same & Median drop & $\geq$2 pp & Median n \\
\midrule
\textbf{Overall} & 20428 & 61.9 & 38.1 & 7.98 & 81.58 & 142 \\
Excluding threshold optimizer & 6345 & 36.3 & 63.7 & 1.95 & 48.24 & 146 \\
Feature mixing & 1232 & 32.1 & 67.9 & 1.37 & 29.46 & 144 \\
Reweighing & 1159 & 25.6 & \textbf{74.4} & 1.36 & 41.50 & 100 \\
Protected-group mixing & 3954 & 40.7 & 59.3 & 2.33 & 56.07 & \textbf{170} \\
\textbf{Threshold optimizer} & 14083 & \textbf{73.4} & 26.6 & \textbf{12.34} & \textbf{96.60} & 134 \\
\bottomrule
\end{tabular}
\endgroup

\end{table}
\appendixpagegap

\FloatBarrier
\section{Operating-point sensitivity of the worst-cell result}
\label{app:operating-point}

\paragraph{Why the operating points differ.}
Threshold optimization is fitted with a balanced-accuracy objective under an equalized-odds
constraint, while every other candidate predicts at its base learner's default threshold. The two are
therefore compared at different operating points, and plain accuracy is sensitive to that
difference. Averaged over all runs, the post-processor loses 13.3 \pp{} of intersectional worst-group
accuracy but only 3.2 \pp{} of intersectional worst-group \emph{balanced} accuracy. Much of the
headline worst-cell drop is thus a shift in operating point, not a real loss to that cell.

\autoref{tab:operating-point} recomputes the complete hidden-regression screen with worst-group
balanced accuracy substituted for worst-group accuracy in \autoref{eq:hidden}, where the balanced
accuracy of a group is the mean of its true-positive and true-negative rates. Weighting the two
classes equally removes the advantage that plain accuracy gives to matching the base rate, and that is
what keeps the comparison stable when the decision threshold shifts. The single-axis screen, the gap
criterion, and both thresholds are unchanged.
The ERM seed-pair null is recomputed under the same substitution so the comparison stays matched.
Every rate falls, the threshold-optimizer rate falls furthest in absolute terms, and the median
worst-cell drop roughly halves. The ordering is unchanged. Threshold optimization remains the only
method far above its null, non-threshold mitigations remain below it, and the qualitative claim in
\S\ref{sec:hidden} holds on either definition. We regard the balanced-accuracy column as the
conservative reading of the result.

\begin{table}[!htbp]
  \caption{Hidden intersectional regression under two definitions of the worst-cell criterion. WGA is worst-group accuracy, as used in the main text; WGBA is worst-group balanced accuracy, which is invariant to the operating-point difference between the post-processor and the other candidates. Rates are over single-axis improvement cases; median drops are conditional on the corresponding flag firing.}
  \label{tab:operating-point}
  \centering
  \small
  \setlength{\tabcolsep}{5.5pt}
  \renewcommand{\arraystretch}{1.08}
  \begin{tabular}{@{}lrrrrrr@{}}
    \toprule
    & \multicolumn{2}{c}{Hidden (\%)} & \multicolumn{2}{c}{Flag fires (\%)} & \multicolumn{2}{c}{Median drop (pp)} \\
    \cmidrule(lr){2-3}\cmidrule(lr){4-5}\cmidrule(lr){6-7}
    Comparison & WGA & WGBA & WGA & WGBA & WGA & WGBA \\
    \midrule
    ERM seed-pair null & 55.0 & 50.2 & 41.7 & 38.7 & n/a & n/a \\
    Overall & 69.9 & 56.5 & 65.6 & 50.5 & 7.98 & 4.17 \\
    Excluding threshold optimizer & 46.9 & 39.3 & 39.6 & 32.4 & 1.95 & 1.84 \\
    Feature mixing & 36.2 & 40.5 & 25.5 & 32.3 & 1.37 & 1.47 \\
    Reweighing & 29.6 & 26.0 & 26.7 & 23.4 & 1.36 & 1.52 \\
    Protected-group mixing & 65.4 & 46.8 & 57.9 & 38.2 & 2.33 & 2.22 \\
    Threshold optimizer & \textbf{94.2} & \textbf{74.6} & \textbf{93.0} & \textbf{69.7} & \textbf{12.34} & \textbf{6.01} \\
    \bottomrule
  \end{tabular}
\end{table}
\appendixtablegap

\autoref{tab:concentration} reports the distribution behind the levelling-down claim in
\S\ref{sec:hidden}. For every row on which the worst-cell flag fires, concentration is the worst
cell's drop minus the population's drop, so a positive value means the worst-off cell absorbed more
than its share and a negative value means it absorbed less. Threshold optimization is the only method
whose concentration is negative on the operating-point-invariant metric.

\begin{table}[!htbp]
  \caption{Concentration of the loss on flagged rows: worst-cell drop minus population drop, in percentage points. Positive values mean the worst-off intersectional cell absorbs more than the population; negative values mean it absorbs less. The final column gives the share of flagged rows with positive concentration on the balanced-accuracy measure.}
  \label{tab:concentration}
  \centering
  \small
  \setlength{\tabcolsep}{5.5pt}
  \renewcommand{\arraystretch}{1.08}
  \begin{tabular}{@{}lrrrrr@{}}
    \toprule
    & & \multicolumn{2}{c}{Plain accuracy} & \multicolumn{2}{c}{Balanced accuracy} \\
    \cmidrule(lr){3-4}\cmidrule(lr){5-6}
    Method & Flagged rows & Median & Mean & Median & Positive (\%) \\
    \midrule
    Feature mixing & 1{,}232 & +1.44 & +2.08 & +0.80 & 61.6 \\
    Reweighing & 1{,}159 & +1.13 & +2.48 & +0.12 & 53.9 \\
    Protected-group mixing & 3{,}954 & +0.98 & +1.76 & $-0.04$ & 49.3 \\
    Threshold optimizer & 14{,}083 & +0.63 & +0.96 & $\mathbf{-3.56}$ & \textbf{26.0} \\
    \bottomrule
  \end{tabular}
\end{table}
\appendixpagegap

\begin{table}[!htbp]
  \caption{Cluster-bootstrap intervals for the rates that carry \S\ref{sec:hidden} after calibration, resampled over task/split/seed clusters as elsewhere in the paper. The eligibility-filtered rate cannot be separated from its baseline; the worst-cell identity excess can.}
  \label{tab:intervals}
  \centering
  \small
  \setlength{\tabcolsep}{5pt}
  \renewcommand{\arraystretch}{1.08}
  \begin{tabular}{@{}p{0.46\linewidth}rrrl@{}}
    \toprule
    Quantity & Rate (\%) & 95\% CI & $n$ & Baseline \\
    \midrule
    Hidden regression, threshold optimizer, accuracy-eligible rows & 61.7 & 53.5--70.1 & 256 & 55.0, inside CI \\
    Worst-cell identity change, threshold optimizer & 73.4 & 71.2--75.7 & 14{,}083 & 58.4, outside CI \\
    Worst-cell identity change, ERM seed pairs & 58.4 & 52.8--63.8 & 2{,}240 & n/a \\
    \bottomrule
  \end{tabular}
\end{table}
\appendixpagegap

\FloatBarrier
\section{Random-forest validation}
\label{app:rf}

\paragraph{Setup.}
\autoref{tab:rf-validation} reports the validation run behind \S\ref{sec:rf}. It repeats the whole
audit with a regularized random forest in place of logistic regression, holding the tasks, splits,
missingness scenarios, candidate methods, and selection rules fixed, so any movement between the two
columns is attributable to the base learner.

\paragraph{Selection stability degrades.}
The two families of result behave differently under that swap, which is the reason for running it.
Selection stability degrades: strict flips rise from 37.3\% to 41.1\% overall, and the gap is widest
at high protected-label availability, where the logistic audit flips 20.1\% of the time and the
random-forest audit 30.4\%. Because the random forest has higher variance, it produces more near-ties
among candidates, and the selector resolves more of them differently from one run to the next. So
every result in \S\ref{sec:calibration} that compares a flip rate against a baseline is specific to
the base learner. An auditor who swaps in a different learner should recompute the baseline instead of
reusing ours.

\paragraph{The subgroup result holds.}
The subgroup result does not move in the same way. Hidden intersectional regression is 69.4\% against
69.9\%, and the threshold-optimizer figure is 93.5\% against 94.2\%. Excluding threshold optimization,
the rate stays below the seed null at 43.5\% against 46.9\%. The mechanism from \S\ref{sec:hidden} comes
from what the equalized-odds post-processor does to the decision threshold, and this is largely the
same whatever model supplies the scores.

\begin{table}[!htbp]
  \caption{Random-forest validation; entries are percentages. Selection stability degrades with the higher-variance base learner, while the hidden-regression pattern and its concentration in threshold optimization are preserved.}
  \label{tab:rf-validation}
  \centering
  \small
  \setlength{\tabcolsep}{8pt}
  \renewcommand{\arraystretch}{1.05}
  \begin{tabular}{@{}lrr@{}}
    \toprule
    Quantity & Logistic & Random forest \\
    \midrule
    Strict conclusion flips & 37.3 & 41.1 \\
    No-label MCAR flips & 60.2 & 61.9 \\
    80\% MCAR flips & 20.1 & 30.4 \\
    Hidden intersectional regression & 69.9 & 69.4 \\
    Hidden regression excl. threshold optimizer & 46.9 & 43.5 \\
    Threshold optimizer hidden regression & 94.2 & 93.5 \\
    \bottomrule
  \end{tabular}
\end{table}
\appendixpagegap

\FloatBarrier
\section{Missingness mechanisms and subgroup support}
\label{app:missingness}

\begin{figure}[!htbp]
  \centering
  \makebox[\linewidth][c]{%
    \includegraphics[width=1.08\linewidth]{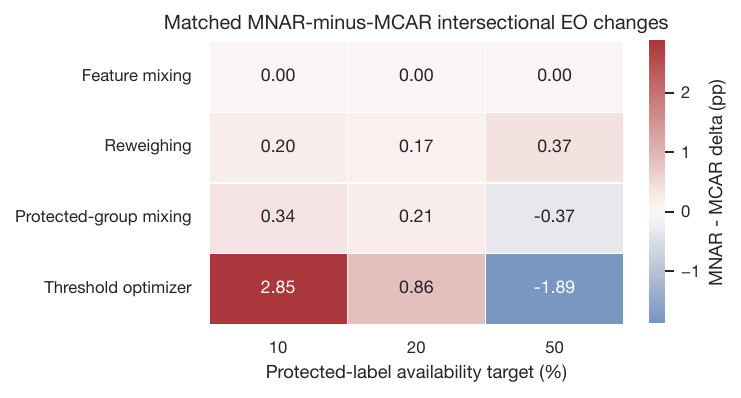}}
  \caption{Matched MNAR-minus-MCAR changes in intersectional EO gap at nominal protected-label availability targets. Positive values indicate that MNAR worsens the intersectional EO gap relative to matched MCAR.}
  \label{fig:mnar}
\end{figure}

\paragraph{Verifying the scenarios.}
Before interpreting availability curves, we verify the simulated label-access scenarios behind them.
This includes realized availability (\autoref{tab:observed-availability}), retained subgroup support
(\autoref{tab:group-retention}), and matched MNAR-minus-MCAR cells (\autoref{fig:mnar} and
\autoref{tab:mcar-vs-mnar}), so the mechanism comparisons rest on the intended missingness scenarios.
These checks are the evidence base for fields 5 and 6 of the \ADS{}.

\begin{table}[!htbp]
  \caption{Nominal target versus realized observed protected-label availability. MCAR tracks the target directly; MNAR uses the majority-versus-rest multiplier, so realized fractions can differ from the displayed x-axis targets.}
  \label{tab:observed-availability}
  \centering
  \small
  \begingroup
\setlength{\tabcolsep}{7pt}
\renewcommand{\arraystretch}{1.12}
\begin{tabular}{@{}lrrrr@{}}
\toprule
Mechanism & Target (\%) & Train observed (\%) & Prediction observed (\%) & n \\
\midrule
MCAR & 0 & \textbf{0.000} & \textbf{0.000} & 2800 \\
MCAR & 10 & 10.000 & 10.000 & 2800 \\
MCAR & 20 & 20.000 & 20.000 & 2800 \\
MCAR & 40 & 40.000 & 40.000 & 2800 \\
MCAR & 50 & 50.000 & 50.000 & 2800 \\
MCAR & 60 & 60.000 & 60.000 & 2800 \\
MCAR & 80 & 80.000 & 80.000 & 2800 \\
MCAR & 100 & \textbf{100.000} & \textbf{100.000} & 2800 \\
\textbf{MNAR} & 10 & 10.010 & 9.968 & 2800 \\
\textbf{MNAR} & 20 & 20.064 & 20.032 & 2800 \\
\textbf{MNAR} & 50 & 50.031 & 49.608 & 2800 \\
\bottomrule
\end{tabular}
\endgroup

\end{table}
\appendixtablegap

\begin{table}[!htbp]
  \caption{Protected-group cells retained after the 25-row retention rule. Possible groups below 25 is computed relative to the nine \RAC{} codes, two \SEX{} codes, and eighteen possible race-by-sex cells.}
  \label{tab:group-retention}
  \centering
  \small
  \begingroup
\setlength{\tabcolsep}{3.8pt}
\renewcommand{\arraystretch}{1.12}
\begin{tabular}{@{}lrrrrrrr@{}}
\toprule
Attribute & Possible & Mean retained & Median & Range & Mean below 25 & Any below 25 (\%) & n \\
\midrule
RAC1P & 9 & 6.6 & 6 & 5-9 & 2.4 & 85.7 & 30800 \\
SEX & 2 & 2.0 & 2 & 2-2 & 0.0 & 0.0 & 30800 \\
\textbf{RAC1P x SEX} & \textbf{18} & \textbf{12.2} & \textbf{12} & \textbf{10-18} & \textbf{5.8} & \textbf{91.4} & \textbf{30800} \\
\bottomrule
\end{tabular}
\endgroup

\end{table}
\appendixpagegap

\appendixpagegap

\begin{table}[!htbp]
  \caption{Matched MNAR-minus-MCAR deltas by method, availability, and metric. Values provide the numeric companion to \autoref{fig:mnar}; positive EO-gap deltas are worse, while positive accuracy and WGA deltas are better.}
  \label{tab:mcar-vs-mnar}
  \centering
  \small
  \setlength{\tabcolsep}{5pt}
  \renewcommand{\arraystretch}{1.14}
  \makebox[\linewidth][c]{%
    \resizebox{1.04\linewidth}{!}{\begin{tabular}{llll}
\toprule
Method & Availability (\%) & Metric & MNAR - MCAR delta (pp) \\
\midrule
ERM & 10 & Accuracy & \textbf{0.0} \\
ERM & 10 & Intersectional EO gap & 0.0 \\
ERM & 10 & Intersectional worst-group acc. & 0.0 \\
ERM & 20 & Accuracy & 0.0 \\
ERM & 20 & Intersectional EO gap & 0.0 \\
ERM & 20 & Intersectional worst-group acc. & 0.0 \\
ERM & 50 & Accuracy & 0.0 \\
ERM & 50 & Intersectional EO gap & 0.0 \\
ERM & 50 & Intersectional worst-group acc. & 0.0 \\
Feature mixing & 10 & Accuracy & \textbf{0.0} \\
Feature mixing & 10 & Intersectional EO gap & 0.0 \\
Feature mixing & 10 & Intersectional worst-group acc. & 0.0 \\
Feature mixing & 20 & Accuracy & 0.0 \\
Feature mixing & 20 & Intersectional EO gap & 0.0 \\
Feature mixing & 20 & Intersectional worst-group acc. & 0.0 \\
Feature mixing & 50 & Accuracy & 0.0 \\
Feature mixing & 50 & Intersectional EO gap & 0.0 \\
Feature mixing & 50 & Intersectional worst-group acc. & 0.0 \\
Protected-group mixing & 10 & Accuracy & -0.3 \\
Protected-group mixing & 10 & Intersectional EO gap & 0.3 \\
Protected-group mixing & 10 & Intersectional worst-group acc. & -0.2 \\
Protected-group mixing & 20 & Accuracy & -0.1 \\
Protected-group mixing & 20 & Intersectional EO gap & 0.2 \\
Protected-group mixing & 20 & Intersectional worst-group acc. & 0.0 \\
Protected-group mixing & 50 & Accuracy & -0.1 \\
Protected-group mixing & 50 & Intersectional EO gap & \textbf{-0.4} \\
Protected-group mixing & 50 & Intersectional worst-group acc. & -0.3 \\
Reweighing & 10 & Accuracy & -0.0 \\
Reweighing & 10 & Intersectional EO gap & 0.2 \\
Reweighing & 10 & Intersectional worst-group acc. & 0.0 \\
Reweighing & 20 & Accuracy & 0.0 \\
Reweighing & 20 & Intersectional EO gap & 0.2 \\
Reweighing & 20 & Intersectional worst-group acc. & 0.1 \\
Reweighing & 50 & Accuracy & 0.0 \\
Reweighing & 50 & Intersectional EO gap & \textbf{0.4} \\
Reweighing & 50 & Intersectional worst-group acc. & 0.0 \\
\textbf{Threshold optimizer} & 10 & Accuracy & -0.1 \\
\textbf{Threshold optimizer} & 10 & Intersectional EO gap & \textbf{2.9} \\
\textbf{Threshold optimizer} & 10 & Intersectional worst-group acc. & 0.4 \\
\textbf{Threshold optimizer} & 20 & Accuracy & 0.1 \\
\textbf{Threshold optimizer} & 20 & Intersectional EO gap & 0.9 \\
\textbf{Threshold optimizer} & 20 & Intersectional worst-group acc. & -0.1 \\
\textbf{Threshold optimizer} & 50 & Accuracy & 0.1 \\
\textbf{Threshold optimizer} & 50 & Intersectional EO gap & -1.9 \\
\textbf{Threshold optimizer} & 50 & Intersectional worst-group acc. & -0.2 \\
\bottomrule
\end{tabular}
}}
\end{table}
\appendixpagegap

\FloatBarrier
\section{Method and task heterogeneity}
\label{app:heterogeneity}

\begin{figure}[!htbp]
  \centering
  \includegraphics[width=\linewidth]{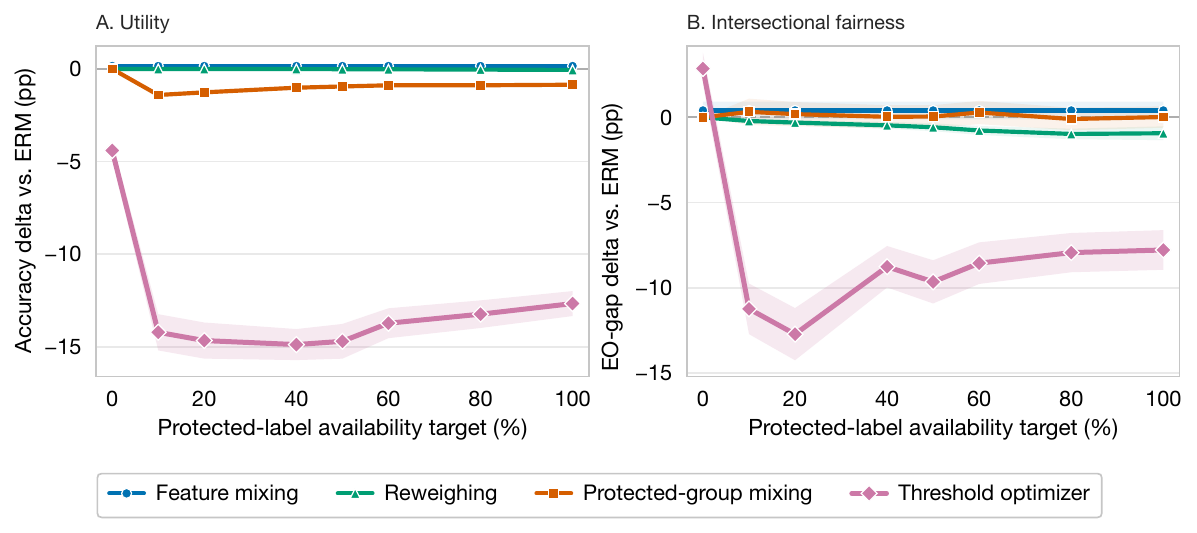}
  \caption{MCAR tradeoff across protected-label availability. Threshold optimization lowers intersectional EO gaps while losing utility and WGA as labels disappear.}
  \label{fig:tradeoff}
\end{figure}

\begin{figure}[!htbp]
  \centering
  \makebox[\linewidth][c]{%
    \includegraphics[width=1.20\linewidth]{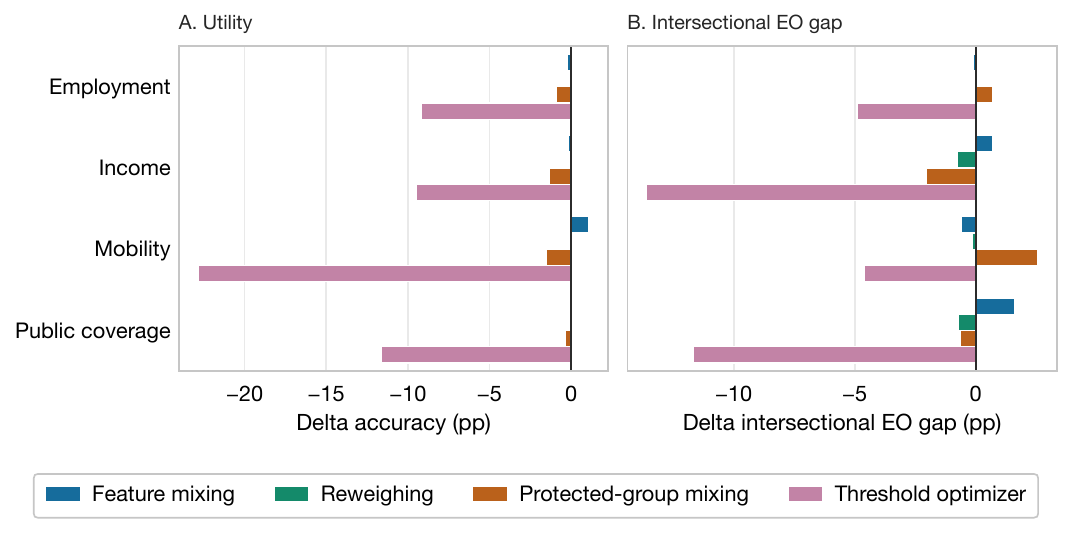}}
  \caption{Task-level method deltas relative to ERM. Effects vary across ACS tasks, motivating per-task as well as aggregate reporting.}
  \label{fig:task}
\end{figure}

\paragraph{Where the tradeoffs come from.}
Aggregate method tradeoffs can come from different metrics or different ACS tasks. This appendix
gives the detail behind \S\ref{sec:tradeoffs}, covering average method behavior
(\autoref{tab:method-deltas-main}) and its expanded form (\autoref{tab:method-deltas-full}), secondary
fairness metrics (\autoref{tab:secondary-fairness}), the availability-curve figure
(\autoref{fig:tradeoff}), and a task-level decomposition (\autoref{fig:task} and
\autoref{tab:task-deltas}).

\begin{table}[!htbp]
  \caption{Average method behavior relative to ERM. Positive accuracy and WGA deltas are better; negative EO-gap deltas are better. Unsafe flags mark cases where aggregate performance is acceptable while subgroup harm worsens.}
  \label{tab:method-deltas-main}
  \centering
  \small
  \begingroup
\setlength{\tabcolsep}{3.2pt}
\renewcommand{\arraystretch}{1.05}
\begin{tabular}{@{}lrrrrr@{}}
\toprule
Method & Acc. $\Delta$ $\uparrow$ & Bal. acc. $\Delta$ $\uparrow$ & Intersec. EO $\Delta$ $\downarrow$ & Intersec. WGA $\Delta$ $\uparrow$ & Unsafe (\%) $\downarrow$ \\
\midrule
Feature mixing & \textbf{+0.160} & -0.050 & +0.408 & +0.096 & \textbf{30.7} \\
Reweighing & -0.012 & -0.059 & \textbf{-0.417} & -0.125 & 1.4 \\
Protected-group mixing & -1.033 & -0.341 & +0.143 & -1.131 & 3.1 \\
Threshold optimizer & \textbf{-13.274} & \textbf{-5.727} & \textbf{-8.685} & \textbf{-13.262} & 1.0 \\
\bottomrule
\end{tabular}
\endgroup

\end{table}
\appendixtablegap

\begin{table}[!htbp]
  \caption{Secondary fairness deltas relative to ERM. Each method cell reports the point estimate and bootstrap confidence interval. Bold marks the largest positive delta in each metric row; positive calibration-Brier and predictive-parity gap deltas are worse. Calibration-Brier gaps are undefined for threshold optimization, which returns a discrete decision instead of a calibrated score, and are marked \textemdash{} in place of a value.}
  \label{tab:secondary-fairness}
  \centering
  \small
  \begingroup
\setlength{\tabcolsep}{4.2pt}
\renewcommand{\arraystretch}{1.2}
\begin{tabular}{@{}lcccc@{}}
\toprule
Metric & Feature & Reweighing & Protected & Threshold \\
\midrule
\multicolumn{5}{@{}l}{\textit{Calibration-Brier gaps}} \\
Intersectional & \begin{tabular}[c]{@{}c@{}}0.538\\[-1pt]{\scriptsize [0.442, 0.649]}\end{tabular} & \begin{tabular}[c]{@{}c@{}}-0.039\\[-1pt]{\scriptsize [-0.076, 0.003]}\end{tabular} & \begin{tabular}[c]{@{}c@{}}\textbf{0.547}\\[-1pt]{\scriptsize [0.426, 0.674]}\end{tabular} & \textemdash \\
Race & \begin{tabular}[c]{@{}c@{}}\textbf{0.371}\\[-1pt]{\scriptsize [0.286, 0.463]}\end{tabular} & \begin{tabular}[c]{@{}c@{}}-0.002\\[-1pt]{\scriptsize [-0.030, 0.035]}\end{tabular} & \begin{tabular}[c]{@{}c@{}}0.368\\[-1pt]{\scriptsize [0.250, 0.502]}\end{tabular} & \textemdash \\
Sex & \begin{tabular}[c]{@{}c@{}}\textbf{0.101}\\[-1pt]{\scriptsize [0.065, 0.137]}\end{tabular} & \begin{tabular}[c]{@{}c@{}}-0.011\\[-1pt]{\scriptsize [-0.026, 0.004]}\end{tabular} & \begin{tabular}[c]{@{}c@{}}0.041\\[-1pt]{\scriptsize [0.005, 0.075]}\end{tabular} & \textemdash \\
\addlinespace[5pt]
\multicolumn{5}{@{}l}{\textit{Predictive-parity gaps}} \\
Intersectional & \begin{tabular}[c]{@{}c@{}}0.036\\[-1pt]{\scriptsize [-0.414, 0.504]}\end{tabular} & \begin{tabular}[c]{@{}c@{}}0.409\\[-1pt]{\scriptsize [0.215, 0.613]}\end{tabular} & \begin{tabular}[c]{@{}c@{}}0.949\\[-1pt]{\scriptsize [0.461, 1.437]}\end{tabular} & \begin{tabular}[c]{@{}c@{}}\textbf{1.078}\\[-1pt]{\scriptsize [-0.086, 2.190]}\end{tabular} \\
Race & \begin{tabular}[c]{@{}c@{}}0.093\\[-1pt]{\scriptsize [-0.229, 0.452]}\end{tabular} & \begin{tabular}[c]{@{}c@{}}0.290\\[-1pt]{\scriptsize [0.148, 0.416]}\end{tabular} & \begin{tabular}[c]{@{}c@{}}0.625\\[-1pt]{\scriptsize [0.207, 1.002]}\end{tabular} & \begin{tabular}[c]{@{}c@{}}\textbf{1.570}\\[-1pt]{\scriptsize [0.636, 2.493]}\end{tabular} \\
Sex & \begin{tabular}[c]{@{}c@{}}0.011\\[-1pt]{\scriptsize [-0.049, 0.068]}\end{tabular} & \begin{tabular}[c]{@{}c@{}}0.198\\[-1pt]{\scriptsize [0.126, 0.264]}\end{tabular} & \begin{tabular}[c]{@{}c@{}}0.227\\[-1pt]{\scriptsize [0.119, 0.339]}\end{tabular} & \begin{tabular}[c]{@{}c@{}}\textbf{0.412}\\[-1pt]{\scriptsize [-0.039, 0.842]}\end{tabular} \\
\bottomrule
\end{tabular}
\endgroup

\end{table}
\appendixpagegap

\begin{table}[!htbp]
  \caption{Expanded method deltas relative to ERM. Rows are metrics and columns are mitigation methods; bold marks the most favorable value in each row. \autoref{tab:method-deltas-main} reports the aggregate subset used in the main results. ROC-AUC is undefined for threshold optimization, which returns a discrete decision instead of a score, and is left blank.}
  \label{tab:method-deltas-full}
  \centering
  \footnotesize
  \begingroup
\setlength{\tabcolsep}{5pt}
\renewcommand{\arraystretch}{1.12}
\begin{tabular}{@{}lrrrr@{}}
\toprule
Metric & Feature mixing & Reweighing & Protected mixing & Threshold opt. \\
\midrule
Accuracy $\Delta$ $\uparrow$ & \textbf{0.160} & -0.012 & -1.033 & -13.274 \\
Balanced acc. $\Delta$ $\uparrow$ & \textbf{-0.050} & -0.059 & -0.341 & -5.727 \\
ROC-AUC $\Delta$ $\uparrow$ & \textbf{0.230} & -0.026 & -0.873 &  \\
\addlinespace[2pt]
Race EO gap $\Delta$ $\downarrow$ & -0.130 & -0.615 & -0.584 & \textbf{-8.169} \\
Sex EO gap $\Delta$ $\downarrow$ & 0.377 & 0.001 & 0.254 & \textbf{-1.712} \\
Intersec. EO gap $\Delta$ $\downarrow$ & 0.408 & -0.417 & 0.143 & \textbf{-8.685} \\
\addlinespace[2pt]
Intersec. WGA $\Delta$ $\uparrow$ & \textbf{0.096} & -0.125 & -1.131 & -13.262 \\
Unsafe flags (\%) $\downarrow$ & 30.7 & 1.4 & 3.1 & \textbf{1.0} \\
\bottomrule
\end{tabular}
\endgroup

\end{table}
\appendixpagegap

\appendixpagegap

\appendixpagegap

\begin{table}[!htbp]
  \caption{Task-level method deltas relative to ERM. Negative EO-gap deltas are better; positive accuracy and WGA deltas are better. Values provide the numeric companion to \autoref{fig:task}.}
  \label{tab:task-deltas}
  \centering
  \small
  \setlength{\tabcolsep}{3pt}
  \renewcommand{\arraystretch}{1.10}
  \makebox[\linewidth][c]{%
    \begin{tabular}{lllll}
\toprule
Method & Task & Delta accuracy (pp) & Delta intersectional EO gap (pp) & Delta intersectional WGA (pp) \\
\midrule
Feature mixing & Employment & -0.2 & -0.1 & -0.2 \\
Feature mixing & Income & -0.2 & 0.7 & \textbf{-0.1} \\
Feature mixing & Mobility & \textbf{1.1} & -0.6 & \textbf{0.9} \\
Feature mixing & Public coverage & -0.1 & 1.6 & \textbf{-0.2} \\
Reweighing & Employment & \textbf{0.0} & -0.1 & \textbf{-0.1} \\
Reweighing & Income & \textbf{-0.0} & -0.8 & -0.1 \\
Reweighing & Mobility & 0.0 & -0.1 & 0.0 \\
Reweighing & Public coverage & \textbf{-0.0} & -0.7 & -0.3 \\
Protected-group mixing & Employment & -0.9 & 0.7 & -0.7 \\
Protected-group mixing & Income & -1.3 & -2.0 & -1.8 \\
Protected-group mixing & Mobility & -1.5 & 2.5 & -1.4 \\
Protected-group mixing & Public coverage & -0.4 & -0.6 & -0.7 \\
\textbf{Threshold optimizer} & Employment & -9.2 & \textbf{-4.9} & -8.5 \\
\textbf{Threshold optimizer} & Income & -9.5 & \textbf{-13.6} & -13.2 \\
\textbf{Threshold optimizer} & Mobility & -22.8 & \textbf{-4.6} & -21.2 \\
\textbf{Threshold optimizer} & Public coverage & -11.6 & \textbf{-11.7} & -10.2 \\
\bottomrule
\end{tabular}
}
\end{table}
\appendixpagegap

\clearpage
\section*{NeurIPS Paper Checklist}

\begin{enumerate}

\item {\bf Claims}
    \item[] Question: Do the main claims made in the abstract and introduction accurately reflect the paper's contributions and scope?
    \item[] Answer: \answerYes{}.
    \item[] Justification: Abstract and introduction claims are scoped to a controlled ACS/Folktables audit with logistic regression as the primary learner and a random-forest validation run. They report seed-null baselines and do not claim that protected-label missingness dominates all audit variation. The Audit Disclosure Standard is presented as a set of reporting fields validated against misreadings observable in this benchmark, not as a policy instrument evaluated with institutional users.

\item {\bf Limitations}
    \item[] Question: Does the paper discuss the limitations of the work performed by the authors?
    \item[] Answer: \answerYes{}.
    \item[] Justification: A dedicated Limitations paragraph names the main scope limits: ACS/Folktables tasks, simulated MCAR/MNAR missingness, this candidate set, four selection criteria, two base-learner families, and the fact that the Audit Disclosure Standard has not been tested with institutional users. The audit setup section and appendix specify row caps, retained-group rules, and the simplified MNAR mechanism.

\item {\bf Theory assumptions and proofs}
    \item[] Question: For each theoretical result, does the paper provide the full set of assumptions and a complete (and correct) proof?
    \item[] Answer: \answerNA{}.
    \item[] Justification: This paper is an empirical audit benchmark and does not present theoretical results or proofs; protocol formulas define evaluation metrics.

    \item {\bf Experimental result reproducibility}
    \item[] Question: Does the paper fully disclose all the information needed to reproduce the main experimental results of the paper to the extent that it affects the main claims and/or conclusions of the paper (regardless of whether the code and data are provided or not)?
    \item[] Answer: \answerYes{}.
    \item[] Justification: Methodology and appendix specify the tasks, years, states, splits, seeds, missingness regimes, candidate methods, metrics, denominators, bootstrap units, row caps, and random-forest validation settings. Supplementary project files include the configuration and artifact-generation workflow used to reproduce the reported figures and tables.

\item {\bf Open access to data and code}
    \item[] Question: Does the paper provide open access to the data and code, with sufficient instructions to faithfully reproduce the main experimental results, as described in supplemental material?
    \item[] Answer: \answerYes{}.
    \item[] Justification: Supplementary project files contain code, configs, workflow scripts, artifact builders, and README commands; raw ACS data are downloaded from public Folktables/Census sources rather than redistributed.

\item {\bf Experimental setting/details}
    \item[] Question: Does the paper specify all the training and test details (e.g., data splits, hyperparameters, how they were chosen, type of optimizer) necessary to understand the results?
    \item[] Answer: \answerYes{}.
    \item[] Justification: Main protocol specifies temporal/geographic splits, seeds, missingness levels, row caps, logistic-regression settings, random-forest validation settings, Fairlearn threshold-optimizer settings, empirical thresholds, and selection criteria.

\item {\bf Experiment statistical significance}
    \item[] Question: Does the paper report error bars suitably and correctly defined or other appropriate information about the statistical significance of the experiments?
    \item[] Answer: \answerYes{}.
    \item[] Justification: Headline rates use 95\% cluster-bootstrap CIs over audit clusters; protocol and appendix state the unit and denominators. Method-delta tables are descriptive aggregates.

\item {\bf Experiments compute resources}
    \item[] Question: For each experiment, does the paper provide sufficient information on the computer resources (type of compute workers, memory, time of execution) needed to reproduce the experiments?
    \item[] Answer: \answerYes{}.
    \item[] Justification: This manuscript reports row caps and single-worker random-forest settings that determine the main memory footprint; supplementary workflow scripts provide the worker resources and regeneration commands for the full artifact build.
    
\item {\bf Code of ethics}
    \item[] Question: Does the research conducted in the paper conform, in every respect, with the NeurIPS Code of Ethics \url{https://neurips.cc/public/EthicsGuidelines}?
    \item[] Answer: \answerYes{}.
    \item[] Justification: This work uses public benchmark data derived from ACS PUMS through Folktables, releases only aggregate/model-output artifacts, and is framed as a diagnostic stress test rather than a deployed decision system.

\item {\bf Broader impacts}
    \item[] Question: Does the paper discuss both potential positive societal impacts and negative societal impacts of the work performed?
    \item[] Answer: \answerYes{}.
    \item[] Justification: The paper is directed at societal impact: it documents that a widely used post-processing mitigation can improve a reported single-axis fairness gap while relocating harm onto an intersectional subgroup, and it proposes reporting fields intended to make that harm visible to oversight readers. Discussion also warns that raw flips, candidate-set near-ties, and single-axis gains can mislead subgroup reliability claims in the opposite direction.
    
\item {\bf Safeguards}
    \item[] Question: Does the paper describe safeguards that have been put in place for responsible release of data or models that have a high risk for misuse (e.g., pre-trained language models, image generators, or scraped datasets)?
    \item[] Answer: \answerNA{}.
    \item[] Justification: This paper does not release high-risk models, scraped data, or individual-level ACS records; the supplementary assets are code, configs, aggregate results, and regeneration scripts.

\item {\bf Licenses for existing assets}
    \item[] Question: Are the creators or original owners of assets (e.g., code, data, models), used in the paper, properly credited and are the license and terms of use explicitly mentioned and properly respected?
    \item[] Answer: \answerYes{}.
    \item[] Justification: Folktables, Fairlearn, and the relevant fairness methods are cited in the paper. Supplementary project documentation records the external packages and data sources used to generate the benchmark artifacts.

\item {\bf New assets}
    \item[] Question: Are new assets introduced in the paper well documented and is the documentation provided alongside the assets?
    \item[] Answer: \answerYes{}.
    \item[] Justification: New assets are benchmark code, configs, workflow scripts, generated aggregate artifacts, and artifact-building scripts. They are documented in the README, configs, appendix tables, and generated artifact files.

\item {\bf Crowdsourcing and research with human subjects}
    \item[] Question: For crowdsourcing experiments and research with human subjects, does the paper include the full text of instructions given to participants and screenshots, if applicable, as well as details about compensation (if any)? 
    \item[] Answer: \answerNA{}.
    \item[] Justification: This paper does not involve crowdsourcing, participant recruitment, intervention, compensation, or direct human-subject interaction.

\item {\bf Institutional review board (IRB) approvals or equivalent for research with human subjects}
    \item[] Question: Does the paper describe potential risks incurred by study participants, whether such risks were disclosed to the subjects, and whether Institutional Review Board (IRB) approvals (or an equivalent approval/review based on the requirements of your country or institution) were obtained?
    \item[] Answer: \answerNA{}.
    \item[] Justification: This work analyzes public benchmark data through Folktables and does not conduct human-subject research or collect new data from participants.

\item {\bf Declaration of LLM usage}
    \item[] Question: Does the paper describe the usage of LLMs if it is an important, original, or non-standard component of the core methods in this research? Note that if the LLM is used only for writing, editing, or formatting purposes and does \emph{not} impact the core methodology, scientific rigor, or originality of the research, declaration is not required.
    \item[] Answer: \answerNA{}.
    \item[] Justification: Core methods, experiments, models, data processing, and evaluation do not use LLMs as important, original, or non-standard components.

\end{enumerate}

\end{document}